\documentclass[10pt,twocolumn,letterpaper]{article}
\pdfoutput=1
\usepackage{iccv}
\usepackage{times}
\usepackage{epsfig}
\usepackage{graphicx}
\usepackage{amsmath}
\usepackage{amssymb}
\usepackage{booktabs}
\usepackage{bbding}
\usepackage{enumitem}
\usepackage{tablefootnote}
\usepackage{enumitem}
\usepackage[accsupp]{axessibility}  % Improves PDF readability for those with disabilities.

\usepackage[pagebackref=true,breaklinks=true,letterpaper=true,colorlinks,bookmarks=false]{hyperref}
\usepackage{authblk}%%llj
\makeatletter
\renewcommand\AB@affilsepx{, \protect\Affilfont}
\makeatother

\iccvfinalcopy % *** Uncomment this line for the final submission

\def\iccvPaperID{9708} % *** Enter the ICCV Paper ID here

\ificcvfinal\fi

\begin{document}

%%%%%%%%% TITLE
\title{RenderIH: A Large-scale Synthetic Dataset for 3D Interacting Hand Pose Estimation}

\author[1,2]{
Lijun Li\thanks{Corresponding author: 4065156@qq.com}}
\author[1]{Linrui Tian}
\author[1]{Xindi Zhang}
\author[1]{Qi Wang}
\author[1]{Bang Zhang} 
\author[3]{Mengyuan Liu}
\author[4]{Chen Chen}

\affil[1]{Alibaba Group}%\vspace{-0.9cm}}
\affil[2]{Shanghai Artificial Intelligence Laboratory}
\affil[3]{Key Laboratory of Machine Perception, Shenzhen Graduate School, Peking University} 
\affil[4]{Center for Research in Computer Vision, University of Central Florida}
%Alibaba Group\\
% Institution1 address\\
%{\tt\small \{\}@alibaba{-}inc.com}
% For a paper whose authors are all at the same institution,
% omit the following lines up until the closing ``}''.
% Additional authors and addresses can be added with ``\and'',
% just like the second author.
% To save space, use either the email address or home page, not both
% \and
% Second Author\\
% Institution2\\
% First line of institution2 address\\
% {\tt\small secondauthor@i2.org}

\maketitle
% Remove page # from the first page of camera-ready.
\ificcvfinal\thispagestyle{empty}\fi

%%%%%%%%% ABSTRACT
\begin{abstract}
The current interacting hand (IH) datasets are relatively simplistic in terms of background and texture, with hand joints being annotated by a machine annotator, which may result in inaccuracies, and the diversity of pose distribution is limited. However, the variability of background, pose distribution, and texture can greatly influence the generalization ability. Therefore, we present a large-scale synthetic dataset --RenderIH-- for interacting hands with accurate and diverse pose annotations. The dataset contains 1M photo-realistic images with varied backgrounds, perspectives, and hand textures. To generate natural and diverse interacting poses, we propose a new pose optimization algorithm. Additionally, for better pose estimation accuracy, we introduce a transformer-based pose estimation network, TransHand, to leverage the correlation between interacting hands and verify the effectiveness of RenderIH in improving results. Our dataset is model-agnostic and can improve more accuracy of any hand pose estimation method in comparison to other real or synthetic datasets. Experiments have shown that pretraining on our synthetic data can significantly decrease the error from 6.76mm to 5.79mm, and our Transhand surpasses contemporary methods. Our dataset and code are available at \textcolor{magenta}{ https://github.com/adwardlee/RenderIH.}

%   Hand interaction is difficult while important since it is essential to better behavior understanding. Most interacting hand models are trained on InterHand2.6M dataset that contain millions of real hand images
% collected from the lab. However, the annotations often contain label noise and the background is simple. We introduce a large-scale synthetic dataset, BlenderHand, for interacting hand pose estimation.
% %by rendering digital hands using a computer graphics pipeline.
% We also propose a simple transformer based network, TransHand, to leverage the correlation between interacting hands. 
% Extensive experiments demonstrate that pretraining on our synthetic data can significantly reduce the error and outperforming contemporary methods. 
\end{abstract}

%%%%%%%%% BODY TEXT
%\vspace{-0.4cm}
\section{Introduction}
\label{sec:intro}

\begin{figure}[t]
	\centering
	\includegraphics[width=0.48\textwidth]{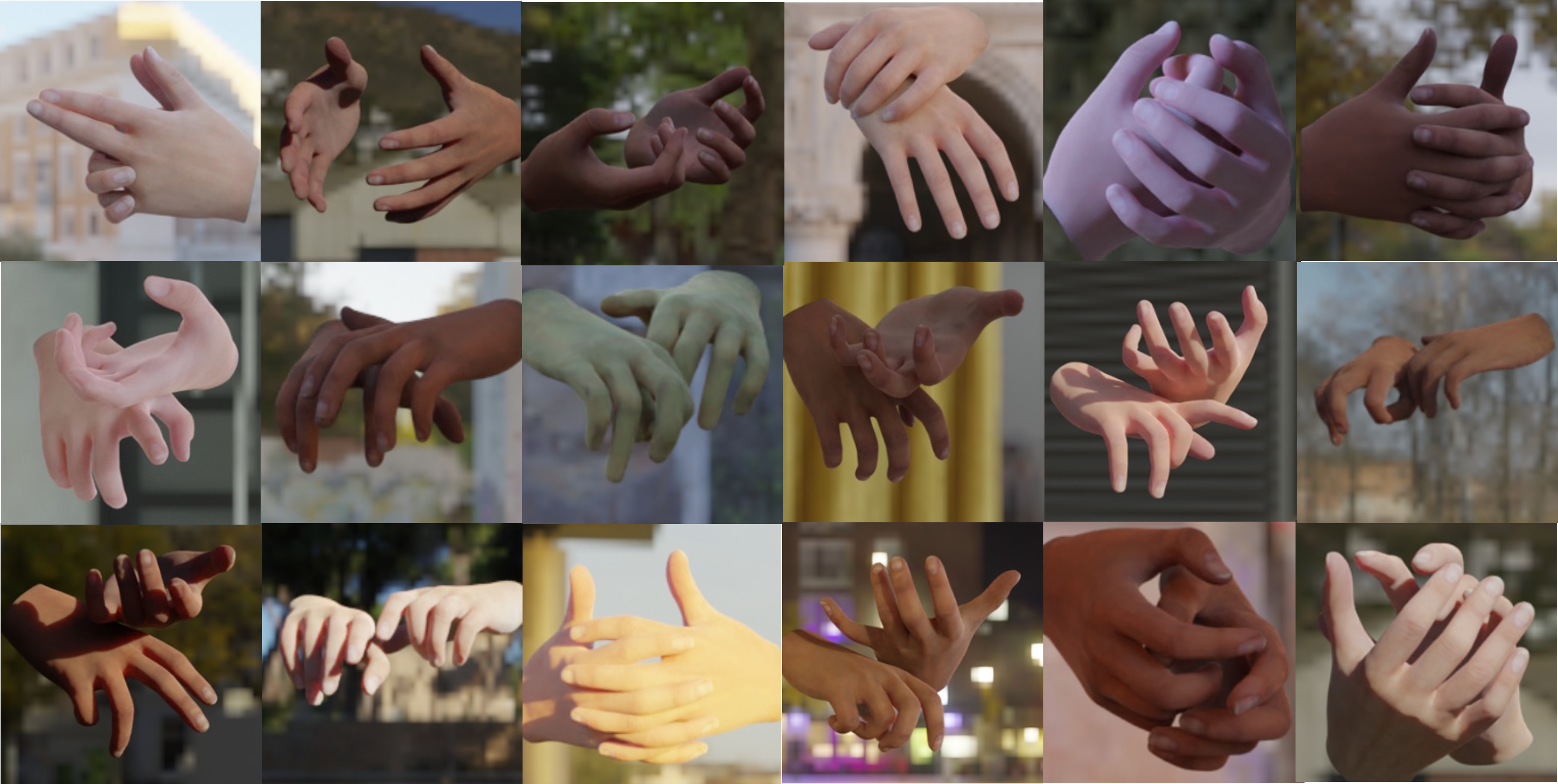}
	\caption{\textbf{Randomly selected samples from RenderIH dataset.} The rendered hands are realistic and varied, capturing a variety of poses, textures, backgrounds, and illuminations.
	}
	\label{example}
\vspace{-0.1cm}
\end{figure} 
% 3D Hand pose estimation is an active research topic as human gesture is a natural way to do human-machine interaction. It has a wide range of applications such as virtual/augmented reality, action/sign language recognition, and gesture control.  However, estimating an accurate 3D hand pose from a monocular RGB image is an ill-posed problem, the depth is hard to estimate. It also remains challenging due to the heavy occlusion, various poses/shapes, different lighting, and background clutter. Spurred by recent development in deep learning techniques, much progress has been made in improving single 3D hand pose estimation from the monocular RGB image ~\cite{graphormer,bmc, minimal, peclr}. 
3D interacting hand (IH) pose estimation from a single RGB image is a key task for human action understanding and has many applications, such as human-computer interaction, augmented and virtual reality, and sign language recognition. However, obtaining 3D interacting hand pose annotations from real images is very challenging and time-consuming due to the severe self-occlusion problem. Some previous works~\cite{h2o3d,cmustudio} have collected some real hand interaction data using a sophisticated multi-view camera system and made manual annotations, but the amount of data is limited. Synthetic 3D annotation data has become increasingly popular among researchers because of its easy acquisition and accurate annotation~\cite{synthhands,ego3d,mvhm, dart, obman,deocclusion, rhd}. 
However, there remain two main challenges: the validity of the generated 3D hand poses and the diversity and realism of the generated images. Therefore, in this paper, we present a high-fidelity synthetic dataset of 3D hand interaction poses for precise monocular hand pose estimation.
% However, one challenge with current synthetic gesture data is that it lacks realism in terms of hand model texture and rendering quality. 

% However, two main limitations exist in the current 3D synthetic hand pose datasets. The first is that they
\begin{table*}[t]
    \small
    \centering
\renewcommand{\arraystretch}{0.85}
\begin{tabular}{l|crcccccc}
   \toprule
    \bf{Dataset}\centering & \bf{Type} & \bf{Data size} &\bf{MT} &\bf{AP} &\bf{background} & \bf{illumination} & \bf{Hand type}& \bf{IH Size}\\
   \midrule
   
   NYU \cite{nyu} & real & 243K & -& \XSolidBrush & lab& uniform & SH & -\\
   STB \cite{stb} & real  & 36K & -& \XSolidBrush & lab& uniform & SH & -\\
   H2O-3D \cite{h2o3d} & real& 76K & -& \XSolidBrush & lab & uniform & HO & -\\
   H2O \cite{h2o} & real& 571K & -& \XSolidBrush & indoor scenes& uniform & HO & -\\
   MVHM \cite{mvhm}& synthetic & 320K & \XSolidBrush & \XSolidBrush& static scenes& uniform & SH & - \\
   ObMan \cite{obman}&synthetic  & 147K& \Checkmark & \XSolidBrush& static scenes& uniform & HO &-\\
   %AIH\cite{amodal} & synthetic & synthetic & 256\times 256 & 3.0M& \XSolidBrush & \bf{IH} & 23\%\\
   DARTset \cite{dart} & synthetic & 800K & \Checkmark& \XSolidBrush& static scenes & manual &SH & -\\
   \midrule
   IH2.6M \cite{interhand26} & real & 2.6M & -& \XSolidBrush & lab & uniform & \bf{IH} & 628K\\
   Ego3d \cite{ego3d} & synthetic & 50K & \XSolidBrush &\XSolidBrush & static scenes& random & \bf{IH} & 40K\\
    \hline
    {\textbf{RenderIH (Ours)}} & synthetic & 1M & \Checkmark & \Checkmark& HDR scenes & \textbf{dynamic} & \bf{IH} & \textbf{1M} \\
   \bottomrule
\end{tabular}
\caption{{\bf{Comparison of the related hand datasets.}} ``MT" is short for multi-textures and means whether the hand models in the dataset are assigned with diverse textures, AP is short for anti-penetration, ``Hand type" means which interaction type the dataset focus on (SH-single hand, HO-hand to object, IH-hand to hand), and ``IH Size" means the proportion of IH poses. ``HDR" is short for High Dynamic Range. Static scenes refer to the use of randomly selected images as the background.}
\label{datasets}
\vspace{-0.2cm}
\end{table*}

Firstly, ensuring the validity of the generated 3D interacting hand poses is a crucial challenge for a synthetic hand system. For example, the pose of Ego3d~\cite{ego3d} is randomized which means a significant portion of the data is not valid. 
To ensure effective hand interactions, the generated two-hand poses must be proximal to each other, while increasing the risk of hand interpenetration.
Therefore, we design an optimization process that considers the constraints of hand attraction and anti-penetration in the meantime, to ensure the proximity of two interacting hands and prevent the occurrence of hand penetration (Section~\ref{subsec:pose-optimization}). 
In addition, the plausibility of hand poses must also be considered. 
Hence, we introduce anatomic pose constraints and apply adversarial learning to ensure that the generated hand poses adhere to anatomical constraints and realism.
Benefiting from pose optimization, our generated dataset contains a rich set of validated two-hand interaction poses as shown in Figure~\ref{example}.

% Hand interpenetration artifacts which are prevalent in existing datasets should be properly addressed. 

Secondly, most existing 3D synthetic hand images lack diversity in terms of backgrounds, lighting, and texture conditions, which prevents them from capturing the complex distribution of real hand data~\cite{ego3d, mvhm, obman}. 
% \textcolor{red}{Todo: take an example of an existing dataset}
% Most of the existing datasets, including Ego3D~\cite{ego3d}, Obman~\cite{obman}, MVHM~\cite{mvhm}, ignore the image quality and the importance of diversity.
% For example, Ego3D~\cite{ego3d} completely adopts the texture of MANO model~\cite{mano} which is quite unrealistic and lack of diversity.  
% On the contrary, we introduce diverse textures, backgrounds, and lighting effects in our rendering system, which can be combined to produce vivid and realistic synthetic hand images (Section~\ref{sssec:ren}). 
% Combined with HDR background, dynamic lighting and ray-tracing renderer, our method obtains 1000K high-quality gesture images. 
Most existing datasets for hand gesture recognition, such as Ego3d~\cite{ego3d}, Obman~\cite{obman}, and MVHM~\cite{mvhm}, do not consider the quality and diversity of the images. For instance, Ego3d~\cite{ego3d} uses the same texture as the MANO model~\cite{mano}, which is unrealistic and monotonous. In contrast, our rendering system introduces various textures, backgrounds, and lighting effects that can produce vivid and realistic synthetic hand images (see Section~\ref{sssec:ren}). By combining HDR background, dynamic lighting, and ray-tracing renderer, we obtain 1M high-quality gesture images (see Figure~\ref{example}).

To assess the performance of our proposed dataset, we carried out comprehensive experiments on it. We demonstrate how much we can reduce the dependency on real data by using our synthetic dataset. Then we contrast our proposed RenderIH with other 3D hand datasets, such as H2O-3D~\cite{h2o3d} and Ego3d~\cite{ego3d}, by training a probing model for each of them and testing on a third-party dataset. Finally, we train a transformer-based network on a mixed dataset of RenderIH and InterHand2.6M (IH2.6M) and achieve state-of-the-art (SOTA) results on 3D interacting hand pose estimation. Our main contributions are as follows:
% \textbf{Our contributions.} To alleviate the issues mentioned above, we propose a large-scale photo-realistic synthetic interacting hand pose dataset, RenderIH which is short for rendering interacting hand. 
% \textbf{Our Contributions.} 
%Note that we prioritize the pose diversity and anti-penetration in RenderIH. 
\begin{itemize}[leftmargin=*,itemsep=2pt,topsep=0pt,parsep=0pt]
\item We propose an optimization method to generate valid and natural hand-interacting poses that are tightly coupled and avoid interpenetration. For image generation, we design a high-quality image synthesis system that combines rich textures, backgrounds, and lighting, which ensures the diversity and realism of the generated images.

\item Based on our data generation system, we construct a large-scale high-fidelity synthetic interacting hand dataset called \textbf{RenderIH}, which contains 1 million synthetic images and 100K interacting hand poses. To the best of our knowledge, this is the largest and most high-quality synthetic interacting dataset so far.

% \item In total, RenderIH contains 1 million synthetic images with various indoor and outdoor background and a larger variety of poses than InterHand2.6M (IH2.6M)~\cite{interhand26}. The advantages of the dataset include the largest number of IH, rich diversity in poses, hand textures, illuminations. The number is one million because our experiments show it make a good balance between training cost and accuracy.

%  \item We propose a new pose optimization algorithm to generate anti-penetration and more natural poses. With theses poses, we design a rendering pipeline to generate synthetic IH images.

\item We conduct extensive experiments to verify the effectiveness of our proposed dataset-RenderIH. The results show that with the help of our synthetic dataset, using only 10\% of real data can achieve comparable accuracy as the models trained on real hand data. We also propose a transformer-based network that leverages our dataset and achieves SOTA results.

\end{itemize}

\section{Related work}
\label{sec:relate}

\subsection{Realistic hand dataset}
Establishing a realistic hand dataset is a tedious and challenging procedure, most realistic data are collected by different sensors~\cite{interhand26, megatrack, ho3d, freihand, stb, cmu,h202} including multiple cameras and depth sensors. STB dataset~\cite{stb} obtained 3D annotations of a single hand (SH) via 2D manual labels and depth data. Since manual annotations are time-consuming~\cite {interhand26}, some researchers~\cite{cmu,interhand26, megatrack, freihand} utilized semi-automatic methods to make annotations. Moon et al.~\cite{interhand26} captured hand interactions with hundreds of cameras. They manually annotated the 2D keypoints of both hands on a few images and utilized a machine detector to help annotate the rest data. While some researchers~\cite{ho3d, contactpose, nyu} proposed automatic methods to make annotations, Hampali et al.~\cite{ho3d} collected hand-object (HO) interactions and jointly optimized 2D key points on multiple RGB-D images to estimate 3D hand poses. Some researchers~\cite{fhb,h2o,mhp} obtain the 3D annotations of hands via some special equipment. Ye et al.~\cite{h2o} captured hand poses via multiple joint trackers. Due to the limitation of the data collection scene, most realistic datasets are in simple scenarios, e.g. lab~\cite{stb, cmu, ho3d} or green screen~\cite{freihand,interhand26,h2o,contactpose, dsp}. Most realistic datasets focus on SH or HO interactions and very few papers~\cite{interhand26, dsp} collect interacting hand data.

\subsection{Synthetic hand dataset}
To obtain precise annotations and increase the dataset's diversity, several papers~\cite{synthhands,ego3d,mvhm, dart, obman,deocclusion, rhd} established synthetic hand dataset by applying multiple backgrounds~\cite{rhd} or different hand textures~\cite{dart}. Most datasets~\cite{mvhm,dart,synthhands,rhd} focus on SH pose data. DARTset~\cite{dart} introduced a shaped wrist and rendered hand images with different skins and accessories. 
%They interpolated the rotations of the fingers to get massive poses, 
But the dataset did not contain IH. To simulate the HO interactions, Hasson et al.~\cite{obman} utilized physics engine~\cite{graspit} to generate object manipulation poses, but their rendered images are not photo-realistic. Although some datasets~\cite{ego3d, deocclusion} provide poses of both hands, the rendered images are not natural enough and lack diversity. Those poses of Ego3d~\cite{ego3d} were randomized, which leads to severe interpenetration between hands and the pose is relatively strange. Based entirely on the pose annotations of IH2.6M~\cite{interhand26}, AIH~\cite{deocclusion} produced a synthetic interacting hand dataset, but only hand masks were created and other annotations were missing.

We summarize some representative hand datasets and compare them to ours in Table~\ref{datasets}. While most datasets focus on SH or HO interactions, they are deficient in handling mesh collision, maintaining high-quality annotations, and providing pose diversity to some extent.
%While some optimization based works\cite{cpf,artiboost,dsp} are proposed to help reconstructing hand interactions, but they are mainly used for realistic detection. 

%------------------------------------------------------------------------
\begin{figure}
	\centering
	\includegraphics[width=0.35\textwidth]{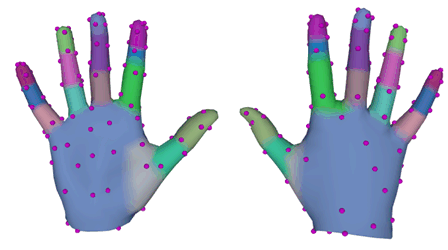}
	\caption{The distribution of anchors and hand subdivision. Purple points denote the anchors.
	}
	\label{anchor}
 \vspace{-0.1cm}
\end{figure} 

\section{RenderIH dataset}
\label{sec:image_gen}

One of the main contributions of our paper is the interacting hand pose optimization method that can generate valid and natural poses. In our paper, {\bf{valid}} poses are non-penetration hands and conform to the anatomic constraint outlined in Table~\ref{angle_limit}. The {\bf{natural}} poses not only conform to the anatomy but also frequently occur in daily life.
We uniformly combine generated poses with a variety of hand textures, high dynamic range (HDR) backgrounds, and camera views. All collections are sampled independently to create images as diverse as possible.  In Section~\ref{subsec:pose-optimization}, we introduce our new hand pose generation algorithm. After hand pose generation, how to render the synthetic image is demonstrated in Section~\ref{sssec:ren}. In Section~\ref{subsec:analysis}, we briefly introduce some statistics about our RenderIH dataset.

\subsection{Interacting hand pose optimization}
\label{subsec:pose-optimization}

 \textbf{Hand model}. Based on the widely used parametric hand model MANO~\cite{mano}, Yang et al.~\cite{cpf} proposed A-MANO, which assigns the twist-splay-bend Cartesian coordinate frame to each joint along the kinematic tree and fit the natural hand better. Therefore, we adopt A-MANO to make our optimization more biologically plausible.  

\begin{figure}
	\centering
	\includegraphics[width=0.45\textwidth]{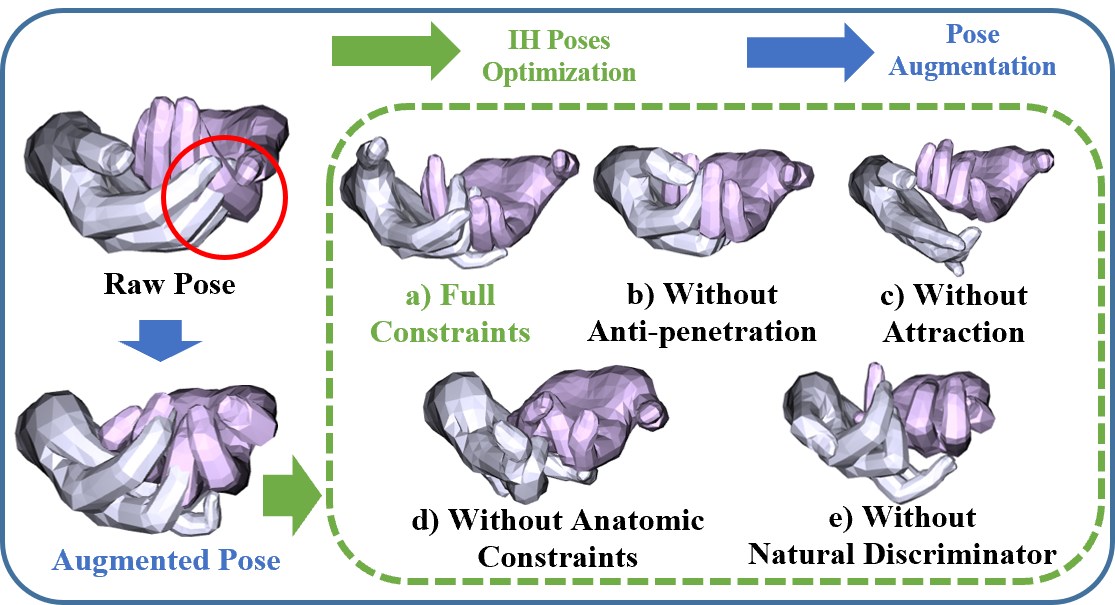}
	\caption{Visualization for the effect of different components in optimization.
	}
	\label{const}
 \vspace{-0.1cm}
\end{figure}

 \textbf{Initial pose generation}. To produce massive valid and natural IH interaction poses, we derive raw poses from the IH2.6M~\cite{interhand26} and then augment the raw poses by assigning random rotation offsets to hand joints. The augmented poses are shown in Figure~\ref{const}, after augmentation, the rotation of the $j_{th}$ finger joint can be expressed as:
 \begin{eqnarray}
 \{R_{ji}\in SO(3)\}_{i=1}^{I}=\{R_{j}R_b(\theta_i^b)R_s(\theta_i^s)\}_{i=1}^{I},
 \end{eqnarray}
where $I$ is the number of augmentation, $R_{b/s}(\theta)$ denotes the rotation along the bend/splay axe, the angle offset $\theta^b\in[-90^{\circ}, 90^{\circ}]$ and $\theta^s\in[-30^{\circ}, 30^{\circ}]$.~
% $R_b(\theta^b)$ and $R_s(\theta^s)$ denote the joint rotates $\theta^b\in[-90^{\circ}, 90^{\circ}]$ and $\theta^s\in[-30^{\circ}, 30^{\circ}]$ along the bend axes and splay axes in respective, the $\theta_i^*$ denotes the $i_{th}$ angle offset.
$SO(3)$ is a group of 3D rotations. $\theta^s=0$ when the joint is not the finger root joint. To avoid abnormal gestures, each augmented joint is restricted according to Table~\ref{angle_limit}. As the augmented poses are totally random, most of them suffer from serious mesh penetration and their gestures are unnatural, it is necessary to optimize the poses.  

 \textbf{Anti-penetration}.
 Inspired by \cite{orisdf}, we adopt multi-person interpenetration loss to interacting hands and propose to divide the hand region into 16 parts. Let $\Omega$ be the modified Signed Distance Field (SDF) ~\cite{modisdf} for each hand. The SDF is defined on a voxel grid of dimensions $N \times N \times N$. It is defined as follows:
  \begin{eqnarray}
  \Omega(x,y,z)=-min(SDF(x,y,z),0),
 \end{eqnarray}
 % $\Omega$ states that its value within a hand is positive and proportional to the distance from the surface, while it is simply zero outside. The penetration loss is calculated as follows:
 %   \begin{eqnarray}
 %  L_p=\sum_{v\in\{V_l\}}\Omega_r(v) + \sum_{v\in\{V_r\}}\Omega_l(v)
 % \end{eqnarray}

 % $V_l$ means left hand vertices and $V_r$ means right hand vertices. While the hand is highly articulated with complex pose and shape, basic hand mesh SDF is not accurate enough. The hand is divided into 16 parts based on its joint position, we compute a separate $\Omega$ function for each hand submesh which is divived according to the hand subdivision in Figure~\ref{anchor}. After applying for each submesh, the penetration loss is defined as:
 %    \begin{eqnarray}
 %  L_p=\sum_{i=1}^{N}\sum_{j=1}^{N}(\sum_{v\in\{M_{lj}\}}\Omega_{ri}(v)) + \sum_{i=1}^{N}\sum_{j=1}^{N}(\sum_{v\in\{M_{rj}\}}\Omega_{li}(v))
 % \end{eqnarray}
where $\Omega$ states that its value within a hand is positive and proportional to the distance from the surface, and it is simply zero outside. The penetration loss for a single hand is calculated as follows:
   \begin{eqnarray}
  L_p^s=\sum_{v\in\{V\}}\Omega_{\Hat{s}}(v).
 \end{eqnarray}
 $V$ means the hand vertices, $s$ is the side of the hand, and $\Hat{s}$ is the side of the other hand. While the hand is highly articulated with a complex pose and shape, basic hand mesh SDF is not accurate enough. We propose to divide the hand into 16 parts based on its joint position and compute a separate $\Omega$ function for each hand submesh which is divided according to the hand subdivision in Figure~\ref{anchor}. After applying for each submesh, the penetration loss is defined as:
    \begin{eqnarray}
  L_p^s=\sum_{i=1}^{N}\sum_{j=1}^{N}(\sum_{v\in\{M_{sj}\}}\Omega_{{\Hat{s}}i}(v)),
 \end{eqnarray}
where $M_{sj}$ means the $j^{th}$ submesh of the hand. The total loss of this part is $L_p = L_p^{right} + L_p^{left}$. The detailed visualization comparison between basic SDF loss and our penetration loss is shown in the \textcolor{blue}{supplementary material (SM)}.

\textbf{Interhand attraction}.
When the IH is in close contact, severe hand occlusion may occur, making it difficult to make annotations. Additionally, the available close contact data are limited. To address this problem, it is recommended to ensure the IH remains in tight contact.

To create contact between the hands, simply bringing the closest vertices together would suffice. However, to reduce the optimization's time complexity, we adopt anchors to guide the position and pose of both hands. As shown in Figure~\ref{anchor}, to downsample the hand vertices as anchors, we traverse IH2.6M to assess the contact frequency of each vertex with the other hand. We selected the vertices with the highest contact frequency as the initial anchors and proceeded to sample the remaining vertices sequentially. Subsequently, we skip the 2-hop neighbors and then continue to sample the yet-to-be-selected ones. Finally, we obtained 108 anchors.

If anchor $a_j^l$ on the left and the anchor $a_i^r$ on the right hand are the closest, they will establish an anchor pair, and the loss of anchor pairs is defined as:
 \begin{eqnarray}
 L_{ij}^A = \frac{1}{2}k_{ij}{\Delta{d_{ij}}}^2,
 \end{eqnarray}
 where $\Delta{d_{ij}}=||a_i^r - a_j^l||_2$. And $k_{ij}=0.5* cos(\frac{\pi}{s} \Bar{\Delta{d_{ij}}})+0.5$, in which $\Bar{\Delta{d_{ij}}}$ is the initial distance between anchors pair. This definition means the initially close anchors tend to keep in contact. Especially the factor $s$ is set to $0.02m$, and we set $k_{ij}=0$ if $\Bar{\Delta{d_{ij}}} > s$. The anchor pairs connection and $k_{ij}$ will be rebuilt during the optimization to adapt to dynamically changing IH poses. 
 % Meanwhile, to avoid abnormal anchors corresponding, the loss function can only be established when $\Bar{n_i^a}\cdot \Bar{n_j^a} < -0.6$, in which $\Bar{n^a}$ is the initial mesh surface normal vector of the anchor.

 % To reduce optimization time, the anchor pairs connection and also $k_{ij}$ will keep constant, but they will be rebuilt  by the method mentioned above after a specific number of iterations, to adapt to dynamically changing IH poses. 

% By employing anchors instead of vertices, it can maintain the optimization quality while reducing the computation time. The time complexity has been reduced from $O(N^2)$ to $O(M^2)$, where $N$ denotes the vertex nums, $M$ denote anchor nums and $M << N$. 
 
 However, only these constraints cannot keep interacting poses valid with random joint angles, we further introduce anatomic optimization.

\textbf{Anatomic Optimization.}
%we should make some imgs explain about each optimization
The finger comprises joints, namely the Carpometacarpal joint (CMC), the Metacarpophalangeal joint (MP), and the Interphalangeal joint (IP). According to the coordinates systems of A-MANO, each finger has three joints, and we denote them as root (CMC of thumb, MP of the others), middle (MP of thumb, Proximal IP of the others), and end joint (IP of thumb, Distal IP of the others). Each of them theoretically has 3 DOF. We define the hand pose in Figure~\ref{anchor} as the T-pose, where all rotation angles are zero. The constraints are defined as follows:

\begin{table}
\small

\resizebox{\linewidth}{!}{
\begin{tabular}{c|ccc}
   \toprule
      \textrm{finger}$\backslash$\textrm{joint} & \textrm{root (B,S)} & \textrm{middle (B)} & \textrm{end (B)}\\
   %_{\rm{finger}}$\backslash$^{\rm{joint}} & root (B,S) & middle (B) & end (B)\\
   \midrule
   thumb & $[-20, 40]$,$[-30, 30]$ & $[-8, 50]$ & $[-10, 100]$ \\
   index & $[-25, 70]$,$[-25, 15]$ & $[-4, 110]$ & $[-8, 90]$  \\
   middle & $[-25, 80]$,$[-15, 15]$ & $[-7, 100]$ & $[-8, 90]$ \\
    ring & $[-25, 70]$,$[-25, 15]$ & $[-10, 100]$ & $[-8, 90]$ \\
     pinky & $[-22, 70]$,$[-20, 30]$ & $[-8, 90]$ & $[-8, 90]$ \\
   \bottomrule
\end{tabular}}

\caption{{\bf{Joint rotation limitations.}} The values are in degrees. 'B'/'S' denotes whether the joint can bend/splay.}
\label{angle_limit}
\vspace{-0.1cm}
\end{table}

\begin{itemize}[leftmargin=*,itemsep=2pt,topsep=0pt,parsep=0pt]
\item {\bf{Available rotation directions}.} Middle and end joint can only rotate $\theta_i^b$ around the B (Bend) axe, while the root can also rotate  $\theta_i^s$ around S (Splay) axe. Always keep $\theta_i^t=0$ around the T (Twist) axe. 
% No joints except the hand root joint can rotate around T (twist) axe, thus the optimization should keep $\theta_i^t=0$.
\item {\bf{Angle limitations}.} According to hand kinematics~\cite{finger,finger1}, the joint rotation limitations are presented in Table~\ref{angle_limit}.

\end{itemize}
The anatomic optimization objective for each hand is defined as:
 \begin{eqnarray}
 L_a = \sum_{i=1}^{15}\sum_{a\in\{b,s,t\}}(\beta(\theta_i^a))^2,
 \end{eqnarray}
where $\beta(\theta_i^a)=max(\theta_i^a-\hat{\theta_i^a},0)+min(\theta_i^a - \check{\theta_i^a},0)$ is the deviation of the rotation angle from its range, and $\hat{\theta_i^a}$/$\check{\theta_i^a}$ is the max/min value of $\theta_i^a$'s range.

\begin{figure}
	\centering
	\includegraphics[width=0.4\textwidth]{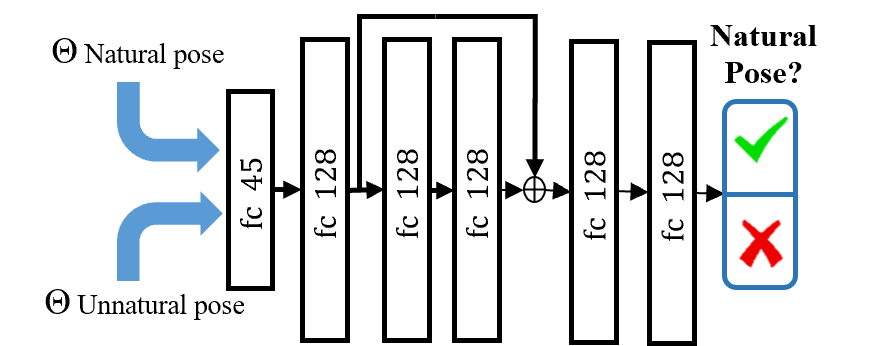}
	\caption{The architecture of the discriminator.
	}
	\label{discriminator}
\end{figure}

\textbf{Natural discriminator.}
After anatomic optimization, the poses become valid. However, as shown in Figure~\ref{const}(e), some optimized poses would not be natural enough. To get the natural poses, we further employ a discriminator $\mathcal{D}$. The detailed structure of the discriminator is illustrated in Figure~\ref{discriminator}. The single-hand pose $\Theta$ is given as input to the multi-layer discriminator. The output layer predicts a value $\in [0,1]$ which represents the probability of belonging to the natural pose. The objective for $\mathcal{D}$ is:
\begin{eqnarray}
    L_{\mathcal{D}}=\mathbb{E}_{\Theta\sim{P_R}}[(\mathcal{D}(\Theta)-1)^2] +\mathbb{E}_{\Theta\sim{P_G}}[\mathcal{D}(\Theta)^2],
\end{eqnarray}
where $P_R$ represents a hand pose from real datasets, such as IH2.6M~\cite{interhand26} and Freihand~\cite{freihand}, $P_G$ is a generated pose. The adversarial loss that is backpropagated to pose optimization is defined as:
\begin{eqnarray}
L_{adv}=\mathbb{E}_{\Theta\sim{P_G}}{[(\mathcal{D}(\Theta)-1)^2]}.
\end{eqnarray}

The discriminator is pre-trained before optimization. We extract 63K natural single-hand poses from Freihand~\cite{freihand}, DexYCB~\cite{dexycb}, and IH2.6M~\cite{interhand26}, their "natural" probabilities $p_n$ are labeled as 1. To get unnatural poses, we follow the methods in "Initial pose generation" to randomly add offsets to the poses, and calculate their probabilities according to the offsets (the higher the offsets, the closer the $p_n$ is to 0). 
%We trained  $\mathcal{D}$ for 5 epochs. 
The qualitative and quantitive improvements brought by $\mathcal{D}$ could be seen in \textcolor{blue}{SM}. Since the natural standard may vary from person to person, we also conducted a user study to confirm the discriminator's effect in \textcolor{blue}{SM}.

%15 joint poses (without root joints, in Euler angles) for each hand are flattened into 1D vectors, and we input the poses of each hand into discriminator to infer whether the hand pose is natural at each iteration of the optimization. We employ the binary softmax loss function $L_n$ to guide the optimization make hands more natural. Refer to supplementary materials for more effects of the discriminator.

\textbf{Poses Optimization.}
In IH optimization, for each hand, it has 15 joints rotation $\Theta = \{R_i\in SO(3)\}_{i=1}^{15}$, hand root rotation $R_r\in SO(3)$ and hand root translation $T_r \in \mathbb{R}^3$, we take $\psi=\{\Theta, T_r\}$ as the optimization parameters and the total IH loss is denoted as:
 \begin{eqnarray}
 \mathop{argmin}\limits_{\psi^r, \psi^l}( w_1\sum_{i=1}^{A_r}\sum_{j=1}^{A_l} L_{ij}^A + w_2 L_a + w_3 L_{adv} + w_4 L_p),
 \end{eqnarray}
where $A_r$/$A_l$ is the anchor numbers of right/left hand, $ L_a=L_a^r + L_a^l$, and $w_*$ is the weight hyperparameter.

% In our implementation, we optimize the loss function in 215 iterations, we assign larger weights for $L_{ik}^R$ in the beginning, and the weights will reduce after 165 iterations when the anchor pairs will also be rebuilt. The learning rate is set to 0.01 and will reduce after 20 no-loss-decaying iterations. Adam solver is utilized for optimization.

%\subsection{Hand texture}
%\label{sssec:texture}

\subsection{Rendering}
\label{sssec:ren}

% Details on how the synthetic dataset is generated are described in the following.
Our dataset offers various benefits, including high-resolution hand textures that create a more natural appearance. Additionally, we simulate natural lighting and environments to address limited diversity in studio settings. Furthermore, our dataset covers a wide range of poses and camera positions, bridging the gap between real-world applications and synthetic data.

\begin{figure}
	\centering
	\includegraphics[width=0.45\textwidth]{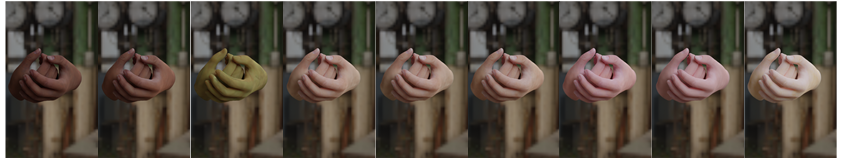}
	\caption{Same hand with different hand textures.
	}
	\label{diff_tex}
 \vspace{-0.1cm}
\end{figure} 

\textbf{Texture.}
To enhance the variety of skin textures we present a broad selection of hues as illustrated in Figure~\ref{diff_tex}.
Color tones include white, light-skinned European, dark-skinned European, Mediterranean or olive, yellow, dark brown, and black. A total of 30 textures are available. In addition, random skin tone parameters can be superimposed on these base skin tones in the shaders to adjust brightness, contrast, and more. Apart from that, these textures also depict wrinkles, bones, veins, and hand hairs to cope with differences in gender, ethnicity, and age.

\begin{figure}
	\centering
	\includegraphics[width=0.45\textwidth]{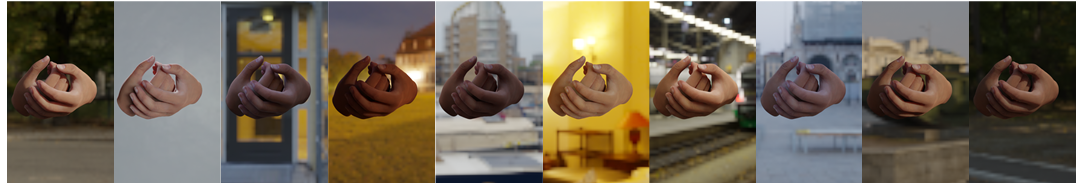}
	\caption{Same hand under diverse illumination.
	}
	\label{hdr}
 \vspace{-0.1cm}
\end{figure} 

\textbf{Lighting and background.} 
It is widely accepted that high-quality synthetic data should resemble real-world scenes as much as possible. 
%For instance, the authors mixed their synthetic human photos with diverse real-world background photographs when creating single-person synthetic data [15].
For instance, the authors mixed their synthetic hands images with diverse real-world background photographs when creating IH synthetic data \cite{ego3d}.
However, simply pasting the rendered hands on the background images is unnatural due to differences in lighting conditions and light angles. 
Since creating a large number of various synthetic 3D background models is time-consuming, we composite synthetic hands with various real-world scenery panoramic images. 
We collect 300 high-dynamic-range (HDR) photography with realistic indoor and outdoor scenes  with appropriate lighting for rendering purposes. They enable our hand models to blend seamlessly with diverse settings resulting in highly photorealistic rendered scenes (see Figure~\ref{hdr}).

\begin{figure}
	\centering
	\includegraphics[width=0.45\textwidth]{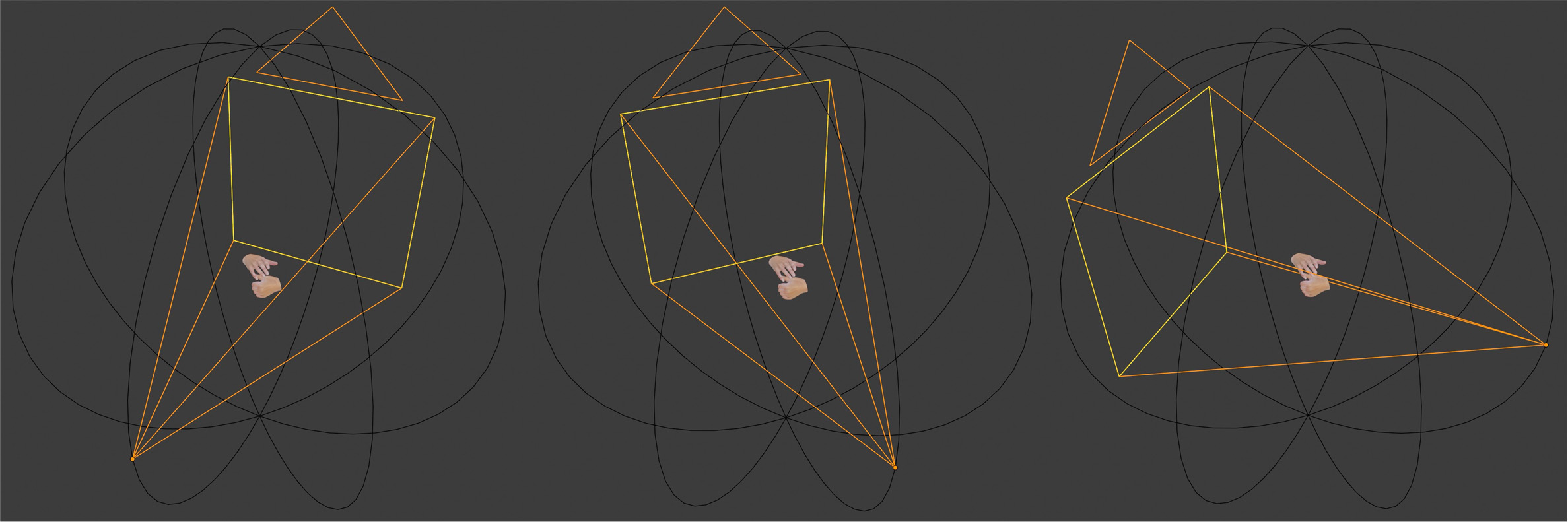}
	\caption{Different viewpoints from the camera track.
	}
	\label{diff_cam}
\end{figure} 

\textbf{Camera Settings.}
%We set up multiple virtual cameras at different locations in the environment to capture the scene from different perspectives.The existing dataset has either a first-person perspective or a third-person perspective. 
We defined a spherical camera arrangement that can contain both viewpoints, enhancing the generalization of the model to different viewpoints.
The center of the two-handed model is first computed and placed at the center of the world, and the camera track is placed around the center with the camera pointing to the center.
Figure~\ref{diff_cam} shows the layout of our simulation environment. 
For each pose, we define four 360-degree circular tracks, which can be averaged by the number of samples to define dense or sparse viewpoints. For sparse sampling, 10 viewpoints were selected for each track.

\begin{figure*}
	\centering
	\includegraphics[width=0.85\textwidth]{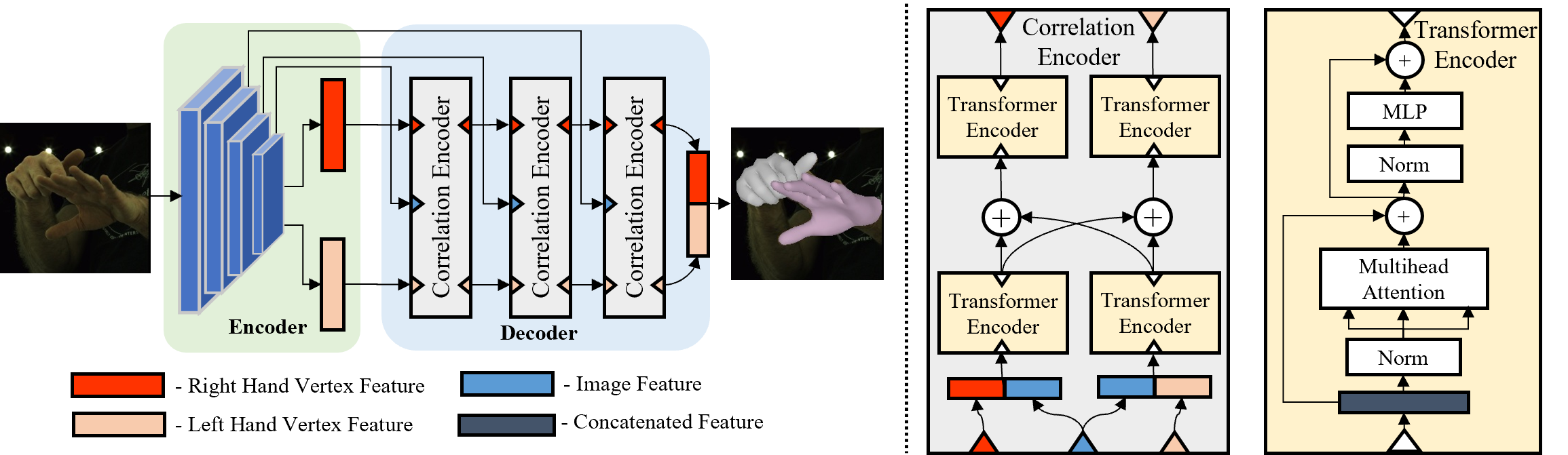}
	\caption{Network architecture. We use the global features extracted by the encoder to predict the left-hand features and right-hand features. After that, our model gradually regresses the hand vertices from 3 identical correlation encoder blocks by fusing multi-resolution image features with hand features. Each correlation encoder contains two transformer encoders and lateral connection from the other hand feature.
	}
	\label{network}
 \vspace{-0.1cm}
\end{figure*}

\begin{figure}[h]
	\centering
	\includegraphics[width=0.45\textwidth]{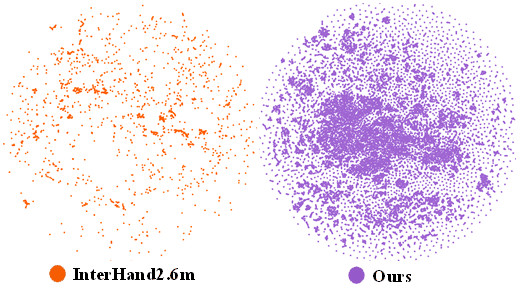}
	\caption{TSNE visualization for IH poses distribution. Our data not only contain the raw poses of IH2.6M but also fill the vacancy by augmentation, resulting in a broader distribution.
	%of InterHand2.6m and ours
	}
	\label{tsne}
 \vspace{-0.1cm}
\end{figure} 

\textbf{Render quality.} 
Our major objective is to improve the photorealism of the synthetic dataset. Therefore, we render the scene in Blender based on the ray-tracing rendering engine Cycles. When creating the hand mesh, we used custom shader settings to adjust the base color, subsurface, and roughness to make the skin more realistic. The resolution of the image is 512$\times$334 pixels and the color depth is 8 bits.

\subsection{Analysis of RenderIH dataset}
\label{subsec:analysis}

%\textbf{Pose Distribution}.
% By collecting annotations from InterHand2.6M, we removed samples of similar poses. At last, only 3792 distinctive poses are left. For each distinctive pose, we augment $I=30$ poses, 40\% of the total are initially in collision (the number of the penetrated mesh faces$>75$, mesh-penetration will be notable if so). After augmenting and optimization, we filter out those IH poses which still have collision between hands (about 93\% of the total remain). Finally we produce 109k natural and non-interpenetration IH poses. From Figure~\ref{tsne}, we can clearly see that our data have much wider pose distribution than IH2.6M. Examples of synthetic images can be seen in Figure~\ref{example}. More statistics details of RenderIH can be seen from supplementary materials.
 For distribution diversity comparison, we project the hand pose in IH2.6M and RenderIH into the embedding space using TSNE~\cite{tsnevis}. Figure~\ref{tsne} clearly shows that our data has a broader pose distribution than IH2.6M. Examples of synthetic images are depicted in Figure~\ref{example} and the rendering video can be found in the \textcolor{blue}{SM}. More visualization effects of different optimization modules and statistical information can be found in the \textcolor{blue}{SM}.

\section{TransHand}
\label{sec:method}
\vspace{-0.1cm}
We propose a transformer-based network, TransHand, for 3D interacting hand pose estimation and conduct extensive experiments on it. 

As the transformer blocks are effective in modeling global interactions among mesh vertices and body joints~\cite{graphormer,maed}, we propose a transformer-based IH network. Our system contains two parts: the encoder and the decoder.  Given an image with size 256$\times$256, the encoder outputs a global feature vector $G_F$ and the intermediate feature maps \{$F_i, i=1,2,3$\} where $i$ indicates the feature level. After that, we map $G_F$ to the left vertex feature $L_F$ and the right vertex feature $R_F$ by using fully connected layers. Since the global feature does not contain fine-grained local details, we concatenate different level features $F_i$ with the hand vertex feature as input to the decoder blocks.

As shown in Figure~\ref{network}, the decoder consists of 3 identical blocks. Each block consists of 2 sub-modules, each sub-module is a typical transformer encoder composed of a multi-head attention module and an MLP layer. Each block is made up of two transformer encoders. As there is usually mutual occlusion in IH, it is natural to combine the other hand feature to improve the estimation precision. Inspired by Slowfast~\cite{slowfast}, we use a symmetric structure to incorporate the other hand feature by adding it, which is the lateral connection in the Correlation Encoder (CE) shown in Figure~\ref{network}. Each block has three inputs including the left vertex feature, right vertex feature, and image feature. The blocks gradually upsample the coarse mesh up to refined mesh and finally to the original dimension with 778 vertices.

 \begin{figure*}[t]
	\centering
    \includegraphics[width=0.85\textwidth]{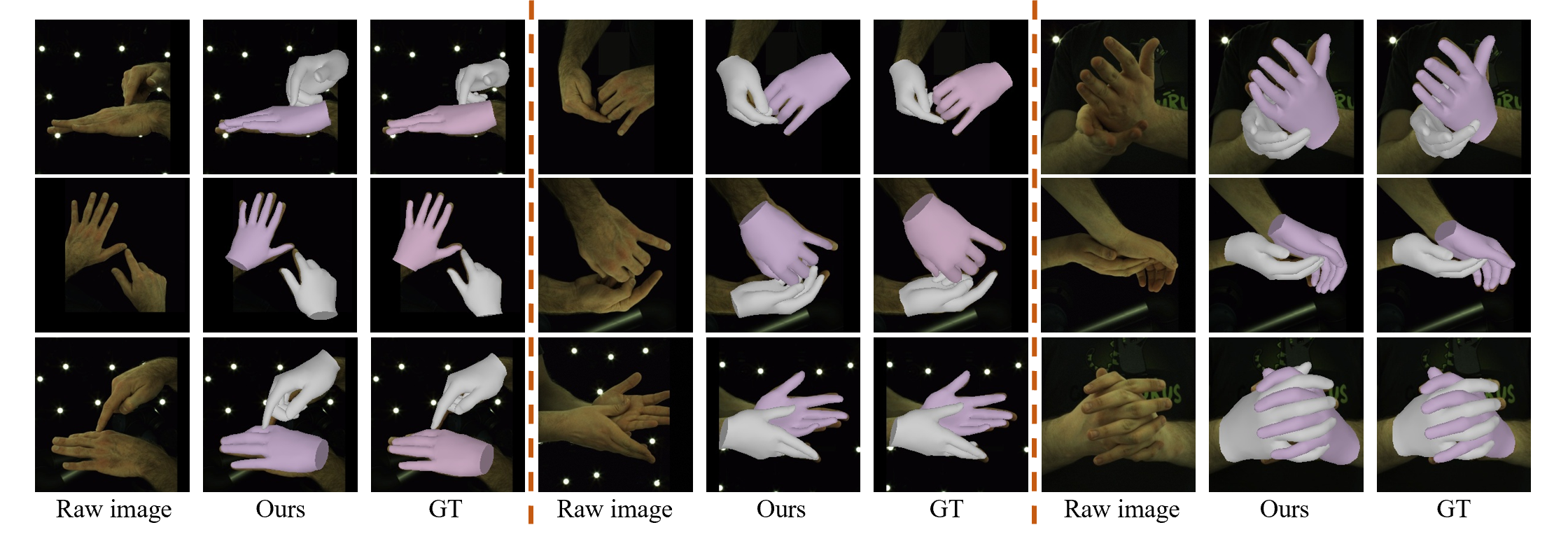}
	\caption{Qualitative results of our method on IH2.6M test set. 
	}
	\label{testset}
\vspace{-0.1cm}
\end{figure*}

\textbf{Loss Function.} For training, we apply $L_1$ loss to 3D mesh vertices and hand joints, and $L_1$ loss to 2D projected vertices and hand joints.
 \begin{eqnarray}
 L_{joint}=\sum_{s=0}^{1}\sum_{i=0}^{M-1}\sum_{d\in\{3D, 2D\}}{\|J_{s,i}^{d}-J_{s,i}^{d,GT}}\|_1, \\
 L_{mesh}=\sum_{s=0}^{1}\sum_{i=0}^{N-1}\sum_{d\in\{3D, 2D\}}{\|V_{s,i}^{d}-V_{s,i}^{d,GT}}\|_1,
 \end{eqnarray}
 where $s$ represents the hand side, $i$ represents the number of joints or vertices, and $d$ denotes whether the computation is for 3D or 2D. To guarantee the geometric continuity of the predicted vertices, smoothness loss is applied which regularizes the consistency of the normal direction between the predicted and the ground truth mesh:
 \begin{eqnarray}
 L_{smooth}=\sum_{s=0}^{1}\sum_{f=0}^{F-1}\sum_{j=0}^{2}{\|e_{f,j,s}\cdot{n_{f,s}^{GT}}\|_1},
 \end{eqnarray}
 where $f$ means the face index of hand mesh, $j$ means the edge of face $f$ and $n^{GT}$is the GT normal vector of this face.
\section{Experiments}
\label{sec:exp}

\subsection{Experiment setup}
\textbf{Dataset}. IH2.6M~\cite{interhand26} is the largest real dataset with interacting hand (IH), and most of our experiments are conducted on this dataset. As we are only focused on IH, we selected only the IH data with both human and machine annotations. After discarding single-hand samples and invalid labeling, we obtain 366K training samples and 261K testing samples. 
Tzionas dataset~\cite{tzionas} is a small IH dataset. We only use it for generalization ability evaluation by using the models trained from different datasets. H2O-3D~\cite{h2o3d} is a real dataset with 3D pose annotations for two hands and an object during interactions. It contains 60K samples. Ego3d~\cite{ego3d} provides 50K synthetic images and corresponding labels of two hands, in which 40K samples are IH and the poses are randomized.

\textbf{Implementation details}. The input images are resized to $256\times256$ and fed to TransHand encoder to generate the global feature and image feature maps. ResNet50~\cite{resnet} is selected as the encoder. For all experiments, the networks are implemented using Pytorch~\cite{pytorch}. We train all models with IH images using Adam optimizer. The initial learning rate is $1e^{-4}$ and the batch size is 64. All experiments are performed on 1 NVIDIA Ampere A100 GPU. 
To demonstrate the usefulness of our RenderIH, we train three mainstream IH pose estimation methods on IH2.6M and a combination of IH2.6M and RenderIH, InterNet\footnote{Since InterNet and DIGIT are trained on the IH subset of IH2.6M v0.0, we train them on v1.0 to make fair comparisons.\label{label1}}~\cite{interhand26}, DIGIT\footref{label1}~\cite{digit} and state-of-the-art method IntagHand\footnote{All the training codes have been open-sourced by the authors.}~\cite{intaghand}.

\textbf{Evaluation metrics}. To evaluate these methods, we report results by two standard metrics: Mean Per Joint Position Error (MPJPE) and the Mean Per Joint Position Error with Procrustes Alignment (PAMPJPE) in millimeters (mm). Additionally, to ensure a fair evaluation with prior research~\cite{intaghand, intershape}, we select the MCP joint of middle finger as root joint and also report SMPJPE which performs scaling to the ground truth bone length. To evaluate the accuracy of estimating the relative position between the left and right hand roots during interaction, we utilize the mean relative-root position error (MRRPE)~\cite{digit} and hand-to-hand contact deviation (CDev)~\cite{arctic} metrics. More results are presented in \textcolor{blue}{SM} with wrist as root joint for future comparison.

\subsection{Results and analysis}
\begin{table}
\centering
\begin{tabular}{cccc}
   \toprule
    With $\mathcal{D}$ & No $\mathcal{D}$ & Raw poses & Augmented poses \\
   \midrule
   81.25\% & 54.68\% & 90.82\% & 32.92\%\\
   \bottomrule
\end{tabular}
\caption{User study on natural rate. The higher the number, the more natural it is.}
\label{natural_test}
\vspace{-0.3cm}
\end{table}

\textbf{User study for naturalness}. Since the perceptions of "natural" may differ from human to human, We conduct experiments to prove the discriminator's effect. We invited 20 persons with/without computer technical background, their ages are from 20 to 60, and the proportion of male to female was approximately 2:1. For each of them, we show 120 pictures (including 30 of augmented poses, 30 of optimized poses, 30 of optimized without discriminator, and 30 of raw poses from IH2.6M) of the IH poses, they are asked to determine whether the shown poses are natural, we count the NR (natural rate) of each category. The results are presented in Table~\ref{natural_test}, the ``Raw poses" are those from IH2.6M\cite{interhand26}, they are performed by humans and have high NR, however, some serious mesh-penetration caused by annotation mistakes might make the testers hardly to determine the ``natural". The ``Augmented poses" are augmented from the raw poses by assigning random rotation offsets to hand joints, they follow the joint limitation but have randomness, and some of them are in mesh penetration, the NR is low in this category. Optimizing the augmented poses without $\mathcal{D}$ solves the penetration, and the poses are valid, but the poses are not natural enough. It is clear that $\mathcal{D}$ improves the naturalness of the poses.

\textbf{Effectiveness of correlation encoder.}
Table~\ref{attention_encoder} shows the performance of models with and without the CE. The baseline method fuses the left-hand feature and right-hand feature with the image feature independently through a transform encoder. The result indicates CE can improve performance by fusing the correlation between hands. Our model is used as default model for subsequent experiments.
\begin{table}[h]
\small
\centering
\begin{tabular}{c|c}
   \toprule
   \text{method}$\backslash$\text{metric} & \text{PAMPJPE/MPJPE/SMPJPE(mm)}$\downarrow$ \\
% _{\rm{method}}$\backslash$^{\rm{metric}} & PAMPJPE/MPJPE/SMPJPE(mm)$\downarrow$ \\
   \midrule
   Baseline & 7.32/11.12/10.82 \\
   Baseline+CE &  6.76/10.6/9.63 \\
   \bottomrule
\end{tabular}
\caption{Effect of correlation encoder (CE) on IH2.6M test set (PAMPJPE/MPJPE/SMPJPE(mm)$\downarrow$). It is shown that CE helps reduce the error by a clear margin.
%to fuse both hand features
}
\label{attention_encoder}
\vspace{-0.1cm}
\end{table}

\begin{table}[h]
\small
\centering
\begin{tabular}{c|cc}
   \toprule
   \textrm{method}$\backslash$\textrm{train set} & 
   IH2.6M & Mixed \\
   % _{\rm{method}}$\backslash$^{\rm{train set}} & IH2.6M & Mixed \\
   \midrule
   InterNet & 18.28 & 17.19  \\
   DIGIT &  15.48 & 14.28   \\
   IntagHand & 10.9  & 9.72\\
   \midrule
   Ours &  10.6 & 10.06  \\
   \bottomrule
\end{tabular}
\caption{Comparison between models trained from IH2.6M and a mixture of RenderIH and IH2.6M in MPJPE(mm)$\downarrow$. The methods are reproduced using their official training code.}
\label{mixed_result}
\vspace{-0.2cm}
\end{table}

\begin{figure}[h]
	\centering
    \includegraphics[width=0.45\textwidth]{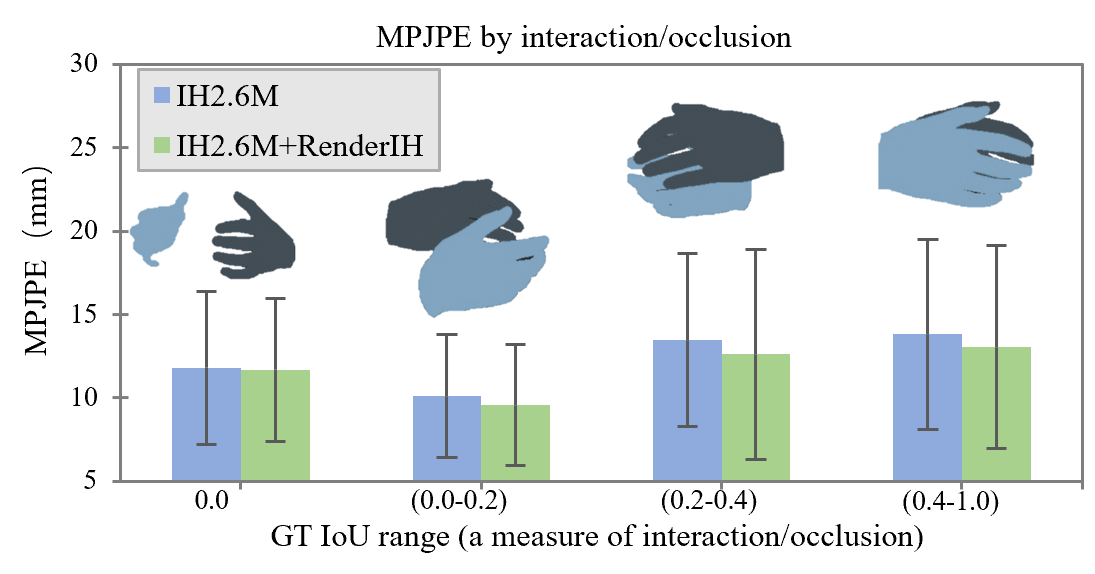}
	\caption{Comparing MPJPE by the degree of occlusion on Tzionas dataset. The IoU between groundtruth left/right masks measures the degree of interaction. The left (yellow) and right
(blue) hand masks provide interaction examples in each IoU range
	}
	\label{tzion_comp}
\vspace{-0.1cm}
\end{figure}

\textbf{Mixing synthetic with real images.}
To demonstrate the usefulness of RenderIH, we test InterNet, DIGIT, IntagHand, and our TransHand on the IH2.6M test set under the setting of training with or without using the full 1M data from the RenderIH dataset. As shown in Table~\ref{mixed_result}, RenderIH is helpful to further reduce the estimation error. For example, the error can be greatly reduced from 10.9mm to 9.72mm for the SOTA IntagHand method. The results prove that our RenderIH has great complementarity with real data. Meanwhile, when hand-hand occlusion is severe, training with our synthetic dataset can handle those cases better than IH2.6M only which is shown in Figure~\ref{tzion_comp}.To quantify the impact of interaction and occlusion, we use the IoU between left and right hand ground truth masks following DIGIT~\cite{digit}. The higher IoU implies more occlusion and half-length of the error bars correspond to 0.5 times of MPJPE standard deviation. With minimal occlusion, the MPJPE is similar between the mixed image model and IH2.6M only. As occlusion increases, the mixed image model reduces MPJPE more substantially than IH2.6M alone. This highlights the value of our RenderIH data.%The bars show the MPJPE over annotated joints for each IoU range. For non-degenerative occlusion (IoU ≤ 0.67), our method has consistent improvement over IH2.6M. In the high IoU regime (> 0.67), the improvement levels off, which is expected since the second hand is almost entirely invisible and the problem is no longer caused by ambiguities. 

\begin{figure}[h]
	\centering
 \includegraphics[width=0.45\textwidth]{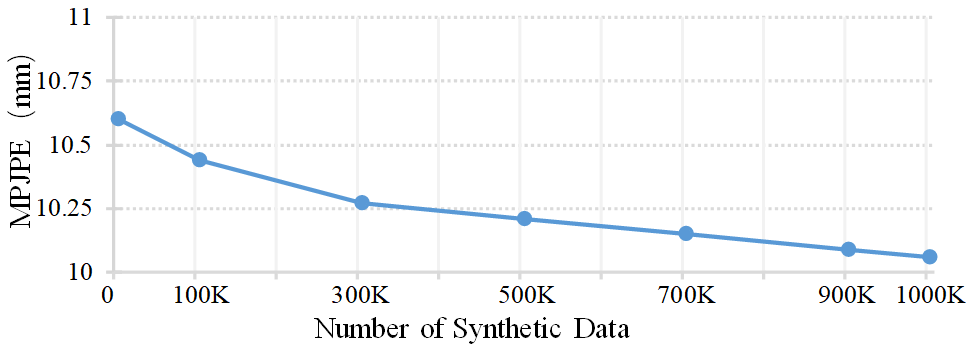}
	\caption{Results of training IH2.6M with different number of RenderIH images on MPJPE(mm)$\downarrow$.
	}
	\label{ablation_syn}
 \vspace{-0.1cm}
\end{figure}

%\subsection{Ablation study}
\textbf{Synthetic data size influence.} During the training phase that involved various combinations of synthetic data and the IH2.6M training set, an obvious decline in the error is observed initially, followed by a gradual decrease after the incorporation of 900K synthetic images, as illustrated in Figure~\ref{ablation_syn}. The trend indicates that beyond a certain volume of synthetic data, the benefits of incorporating additional data become marginal. To balance the cost of training and accuracy, we select 1M as the optimal size for RenderIH.

\begin{figure}[h]
	\centering
 \includegraphics[width=0.45\textwidth]{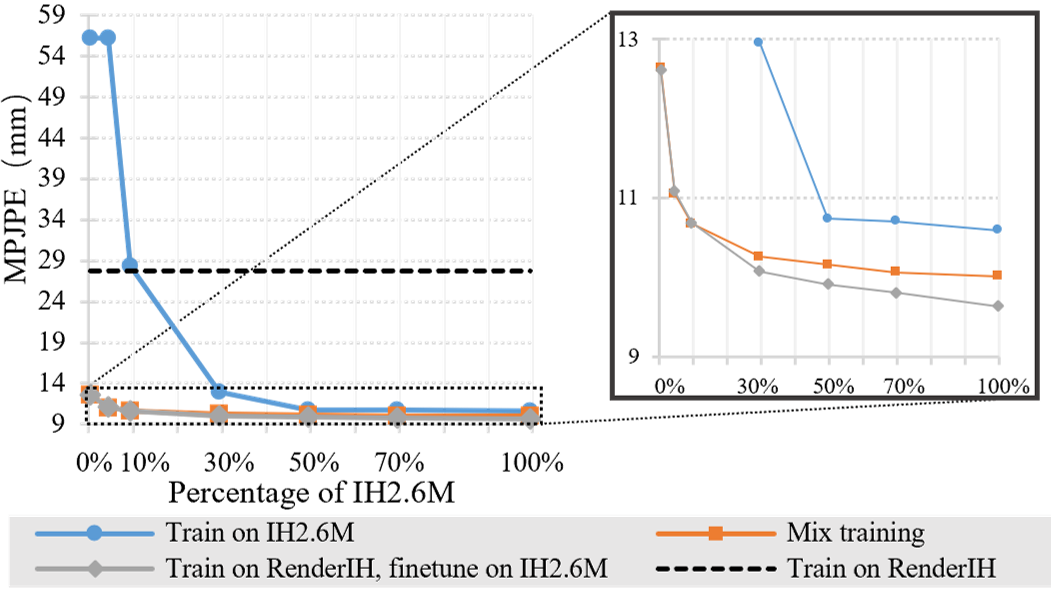}
	\caption{Comparison between training with RenderIH only, with part of IH2.6M only, the combination of the two, with pretraining on RenderIH and finetuning on IH2.6M.
	}
	\label{ablation}
\vspace{-0.1cm}
\end{figure}

\textbf{Training strategy comparison.}
The training strategy of synthetic data and real data is studied in this section. From Figure~\ref{ablation}, both data mix training and pretraining from synthetic data can lead to significantly higher accuracy. Compared to dataset mixing, pretraining on the synthetic followed by fine-tuning on real images led to better precision. In contrast to dataset mixing, our results suggest that pretraining on synthetic data followed by finetuning on real images offers a more effective approach for reducing error.

\begin{table*}[t]
\renewcommand{\arraystretch}{0.9}
\centering
\begin{tabular}{c|ccccc}
   \toprule
   Method & PAMPJPE$\downarrow$ & MPJPE$\downarrow$ & SMPJPE$\downarrow$ & MRRPE$\downarrow$ &CDev$\downarrow$\\
   \midrule
   InterNet$^*$~\cite{interhand26} & 11.72 & 18.28 & 16.68 & - & -\\
   DIGIT$^*$~\cite{digit} & 9.72 & 15.48 & 13.43 & - & -\\
    InterShape~\cite{intershape} & - & - &13.07 & - & -\\
    HDR~\cite{deocclusion} & - & 13.12 & -& - & -\\
    %IntagHand & - & 8.79 \\
    IntagHand~\cite{intaghand} & 6.10 & 10.30 & 8.79 & 12.1 & 25.1\\
    IntagHand{$^*$} & 7.16 & 10.90 & 10.47 & 13.6 & 29.6\\
    \midrule
    Ours & 6.76 & 10.66 & 9.63 & 12.98 & 27.9\\
    Ours{$^\#$} & 5.79 & 9.64 & 8.18 & 11.95 & 24.6\\

   \bottomrule
\end{tabular}
\caption{Comparing with SOTA methods on IH2.6M test set ($*$ means official code reproduction, $\#$ means RenderIH pretraining)}
\label{comparison_sota}
\vspace{-0.1cm}
\end{table*}

\textbf{Real data size influence.}
We study how the real data size affects the estimation precision in Figure~\ref{ablation}. We use all the samples from RenderIH in this section. For real data, we sample the number of data ranging from 3663 to 366358, which takes 1\%, 5\%, 10\%, 30\%, 50\%, 70\%, and 100\% of the real data. Although training only on RenderIH performs poorly, the MPJPE can be greatly reduced from 27.73mm to 12.6mm by finetuning on only 1\% of real data. With finetuning on 10\% of real data, the MPJPE can be almost the same as training on the full real data. When finetuning on all real data, the error can be 0.96mm lower than training only on all real data. 

\begin{table}
\small
\centering
\begin{tabular}{c|cc}
   \toprule
   \textrm{train set}$\backslash$\textrm{test set} &
   IH2.6M & Tzionas\\
% _{\rm{train set}}$\backslash$^{\rm{test set}} & IH2.6M & Tzionas\\
   \midrule
   H2O-3D &  47.81/48.19 & 35.61/34.01 \\
   Ego3d & 58.40/57.48 & 56.90/54.89 \\
   RenderIH & 43.35/42.17 & 32.80/28.70 \\
   \bottomrule
\end{tabular}
\caption{Generalization ability comparison between H2O-3D, Ego3d, and RenderIH on MPJPE/SMPJPE (mm)$\downarrow$. The number of samples is 40K and fixed for each dataset. }
\label{data_general}
\vspace{-0.1cm}
\end{table}

\begin{table}
\small
\centering
\begin{tabular}{c|cc}
   \toprule
   \textrm{train set}$\backslash$\textrm{test set} &
   IH2.6M & Tzionas\\
% _{\rm{train set}}$\backslash$^{\rm{test set}} & IH2.6M & Tzionas\\
   \midrule
   H2O-3D+IH2.6M &  11.05/9.91 & 12.03/12.02 \\
   Ego3d+IH2.6M & 10.66/9.60 & 11.13/11.06 \\
   RenderIH+IH2.6M & 10.58/9.52 & 10.63/10.56 \\
   \bottomrule
\end{tabular}
\caption{Training on the mixture of datasets with all IH2.6M data on MPJPE/SMPJPE (mm)$\downarrow$. The number of samples is 40K for each dataset.}
\label{diffdata}
\vspace{-0.1cm}
\end{table}

% \begin{table}[h]
% \centering
% \small
% \begin{tabular}{l|cc}
%         \toprule
%         Metrics & \multicolumn{2}{c}{MPJPE/MRRPE/CDev$\downarrow$}\\
%          \cline{1-3}
%        _{\rm{Training~set}}$\backslash$^{\rm{Test~set}} & IH2.6M & Tzionas \\
%         \midrule
%                 IH2.6M &  10.6/12.98/27.9 & 11.38/11.1/19.9\\
%                IH2.6M+RenderIH & 9.64/11.94/24.6 & 10.49/9.37/19.5\\
%         \bottomrule
%         \end{tabular}
% \caption{The comparison of training with or without our dataset.}
% \label{renderih}
% \vspace{-0.3cm}
% \end{table}

\textbf{Comparison with H2O-3D dataset, Ego3d dataset and RenderIH subset.} In Table~\ref{data_general} and Table~\ref{diffdata}, we compare the generalization ability of these datasets with the same number of 40K samples. The model pretrained on RenderIH reaches lower error than other models pretrained on H2O-3D and Ego3d in Table~\ref{data_general}, which proves that our artificial data is realistic and the knowledge is more easily transferable. The model trained on RenderIH performs better, possibly because all images have objects that interfere with two-handed interaction in H2O-3D. When training TransHand on RenderIH and IH2.6M, the estimation error is the lowest both in the IH2.6M and Tzionas dataset which is shown in Table~\ref{diffdata}. Especially the result on Tzionas dataset shows our varied pose distribution, background, and texture is helpful for improving generalization.

%\textbf{Comparison between RenderIH and synthetic data generated by original pose of IH2.6M training set.} We can see that model pretrained from RenderIH show much less error on the Interhand testset.

\textbf{Comparison with SOTA methods.}
As is shown in Table~\ref{comparison_sota}, our TransHand can outperform SOTA IntagHand method trained from its official code. Furthermore, their method involves multitask learning and their network comprises of complex graph transformer modules. In comparison, our method is simpler yet highly effective. When pretraining on RenderIH and finetuning on the IH2.6M data, our method can further reduce the MPJPE by about 1mm. Better hand-hand contact (CDev) and better relative root translation (MRRPE) can be observed in this table. Moreover, it is shown in Table~\ref{renderih_comp} that training on our dataset in addition to IH2.6M can lead to obviously lower error on the Tzionas dataset compared with training on IH2.6M alone. Results that are computed with wrist as root is shown in Section 3.3 of \textcolor{blue}{SM}.

\begin{table}[h]
\centering
\small
\begin{tabular}{l|c}
        \toprule
        Metrics & {MPJPE/MRRPE/CDev$\downarrow$}\\
        \cline{1-2}
        \textrm{Training~set}$\backslash$\textrm{Test~set} & Tzionas \\
    % _{\rm{Training~set}}$\backslash$^{\rm{Test~set}} & Tzionas \\
        \midrule
                RenderIH & 22.11/25.8/47.7\\
                IH2.6M &  11.38/11.1/19.9\\
               IH2.6M+RenderIH & 10.49/9.37/19.5\\
        \bottomrule
        \end{tabular}
\caption{The comparison of training with or without our dataset and test on Tzionas dataset.}
\label{renderih_comp}
%\vspace{-0.3cm}
\end{table}
% \iffalse
% \begin{figure}
% 	\centering
%     \includegraphics[0.45\textwidth]{pic/vis.png}
% 	\caption{Qualitative results on in-the-wild image. 
% 	}
% 	\label{inthewild}
%  \vspace{-0.4cm}
% \end{figure}
% \fi

\textbf{Qualitative results}. Our qualitative results are shown in Figure~\ref{testset}. We can see our method can generate high-quality IH results in IH2.6M images. More in-the-wild results can be found in the \textcolor{blue}{SM}.
%Our qualitative results are shown in Figure~\ref{testset}. From Figure~\ref{testset}, we can see our method can generate high quality interacting hand result in InterHand2.6M images. 

%As shown in Figure~\ref{inthewild}, our method can perform well on in-the-wild images. This demonstrates the generalization ability of our network and dataset.
 
%\vspace{-0.2cm}
\section{Conclusion}
\label{sec:conclusion}
In this paper, we propose a new large-scale synthetic dataset for 3D IH pose estimation. Various experiments are conducted to study the effectiveness of RenderIH. With the whole synthetic hand images and only 10\% of real hand images, we can achieve precision that is comparable to the same method which is trained on all the real hand images. We hope that this dataset could be a meaningful step towards developing 3D IH pose estimation models that do not depend on real data and adaptable to to various scenes.

\iffalse
\noindent\textbf{Acknowledgments}. We would like to thank Zhongjian Wang and Ke Sun for their valuable discussion and advice.

\fi
% \vspace{-0.3cm}
% \paragraph{Acknowledgement:} This work was supported in part by

%%%%%%%%% REFERENCES

%\documentclass[10pt,twocolumn,letterpaper]{article}

% \usepackage{iccv}
% \usepackage{times}
% \usepackage{epsfig}
% \usepackage{graphicx}
% \usepackage{amsmath}
% \usepackage{amssymb}
% \usepackage{booktabs}
% \usepackage{bbding}
% \usepackage{enumitem}
% \usepackage{tablefootnote}
% \usepackage{enumitem}
% % Include other packages here, before hyperref.

% %\usepackage[pagebackref=true,breaklinks=true,letterpaper=true,colorlinks,bookmarks=false]{hyperref}
% \renewcommand\thesection{\Alph{section}}
% \iccvfinalcopy % *** Uncomment this line for the final submission
% \usepackage{authblk}%%llj
% \makeatletter
% \renewcommand\AB@affilsepx{, \protect\Affilfont}
% \makeatother

% \def\iccvPaperID{9708} % *** Enter the ICCV Paper ID here
% \def\httilde{\mbox{\tt\raisebox{-.5ex}{\symbol{126}}}}

% % Pages are numbered in submission mode, and unnumbered in camera-ready
% \ificcvfinal\pagestyle{empty}\fi

% \begin{document}
\pagebreak
\begin{center}
\textbf{\large RenderIH: A large-scale synthetic dataset for 3D interacting hand pose estimation (\textit{Supplementary Material})}
\end{center}
%%%%%%%%% TITLE
% \title{RenderIH: A large-scale synthetic dataset for 3D interacting hand pose estimation (\textit{Supplementary Material})}

\maketitle
% Remove page # from the first page of camera-ready.
\ificcvfinal\thispagestyle{empty}\fi
\setcounter{section}{0}

%%%%%%%%% BODY TEXT
This supplementary material contains additional information that could not be included in the main manuscript due to space limitations. We will begin by providing more detailed information about the dataset. Following that, we will briefly discuss the pose optimization details in our approach. Then we will then present additional visualization results from our qualitative experiments. Finally, we will discuss the broader impacts and limitations of our dataset.
\section{More details on RenderIH}
\label{section1}
RenderIH is composed of 1 million synthetic images by varying the pose, camera view, and environment (texture, lighting, and background). By collecting annotations from IH2.6M, we removed samples of similar poses resulting in 3680 distinctive poses. For each distinctive pose, we augment $I=30$ poses. After augmenting and optimization, we filter out those IH poses that still have notable penetration or exceed joint limits, the remaining data accounts for 93\% of the total, and we produce approximately 100K natural and non-interpenetration IH poses. Then we apply 10 camera viewpoints to each pose and produce 1M synthetic images in total. For each image, we randomly pick from a collection of 300 HDR images to illuminate the hand and provide the background together with a hand texture map. The rendering process took more than 200 hours using 4 NVIDIA A100 GPUs.
As for the corresponding annotation, we provide pose and shape parameters, 3D joint coordinates, 2D joint coordinates, and camera intrinsic and extrinsic parameters.
It is worth noting that the synthetic data labels can be freely extended based on the user's preferences, such as generating hand parts segmentation masks. The automatically generated annotations are free of noise and are more flexible than the traditional labels of the real dataset. Some rendering examples to illustrate our photo-realistic effect are provided in the \textbf{video demo}.

\section{Pose optimization details}
\label{section2}
\begin{figure}
	\centering
	\includegraphics[width=0.45\textwidth]{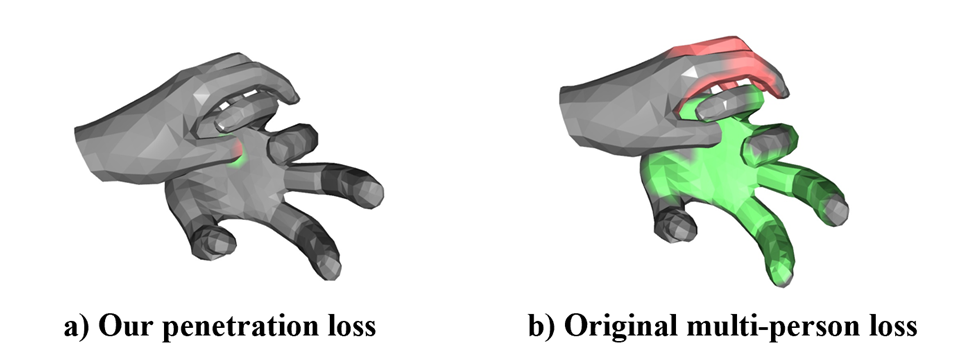}
	\caption{Visual comparison of penetration loss between ours and multi-person penetration loss~\cite{orisdf}. The penetration position of each hand can be obtained by utilizing the penetration loss. The green color and red color are used to denote where the right hand and left hand are penetrated respectively.
	}
	\label{penetration}
\end{figure}

\textbf{Penetration loss comparison.} Hands are highly articulated and have a curved and concave shape as a whole object. It makes the original multi-person penetration loss~\cite{orisdf} hard to detect the correct penetrated positions. However, by dividing the hand into 16 almost convex parts, we can perform SDF on each part and calculate the penetration, which can lead to more accurate penetration detection results. The visual difference between our algorithm and multi-person penetration loss~\cite{orisdf} can be seen from Figure~\ref{penetration}.

\begin{figure}
	\centering
	\includegraphics[width=0.45\textwidth]{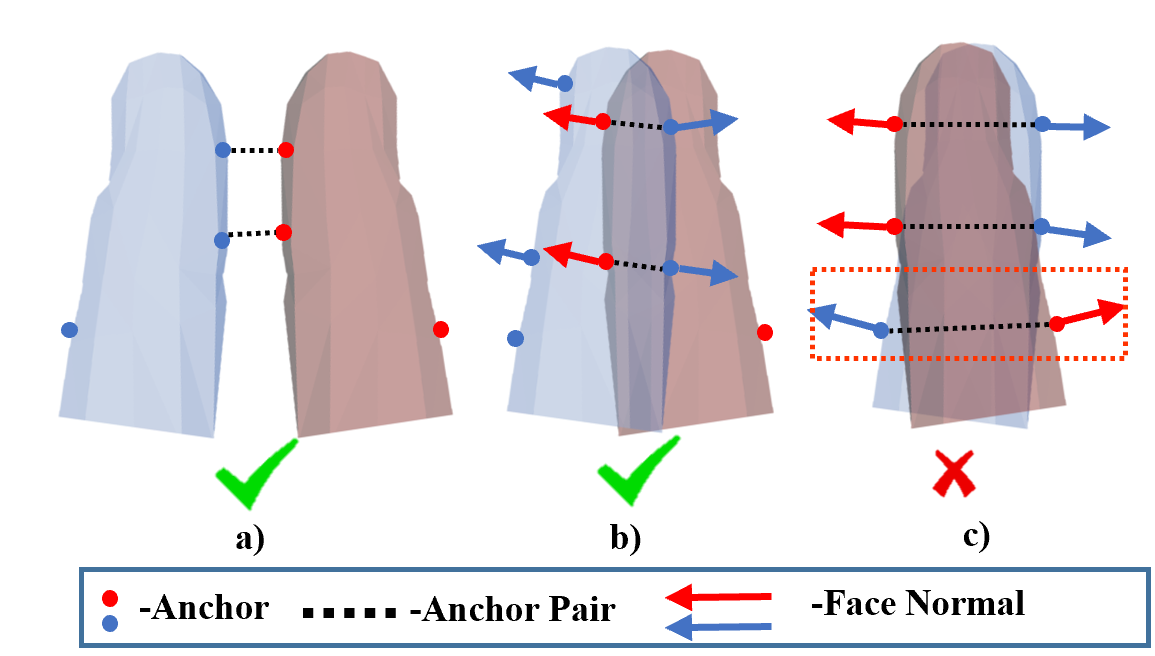}
	\caption{Building anchor pairs between IH. a) Nearest anchors are connected as the pairs, they tend to make more contacts between both hands. b) Using face normals to avoid abnormal pairs. c) The proper pairs are hard to build when the hand parts are in serious overlaps.
	}
	\label{anchor_pair}
\end{figure} 

\begin{figure}
	\centering
	\includegraphics[width=0.5\textwidth]{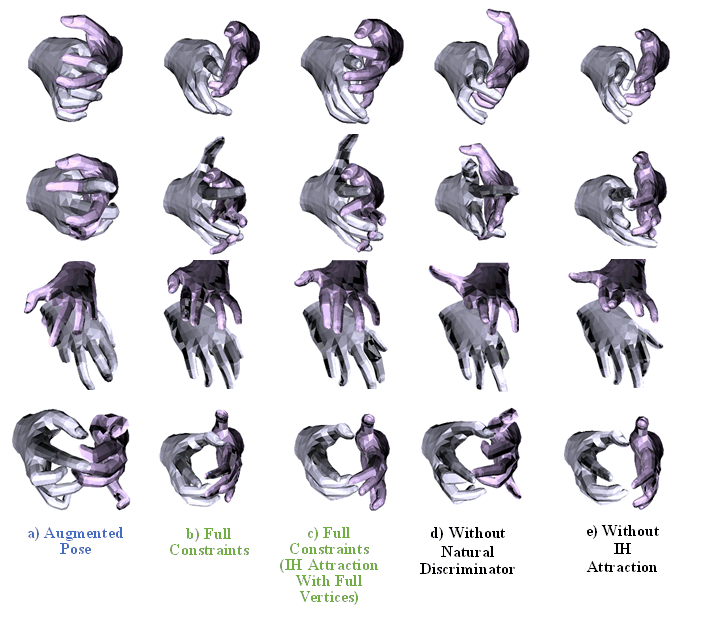}
	\caption{Results for different components of optimization. a) The poses augmented from the IH2.6M. b) After being optimized by the proposed method, the poses become valid and natural. c) Using vertices instead of anchors to make IH attraction has no significant differences. d) Poses optimized without discriminator are valid but not natural. e) Poses optimized without IH attraction have fewer contacts. 
	}
	\label{multiple_pic_const}
\end{figure} 

\begin{figure*}[t]
	\centering
	\includegraphics[width=\textwidth]{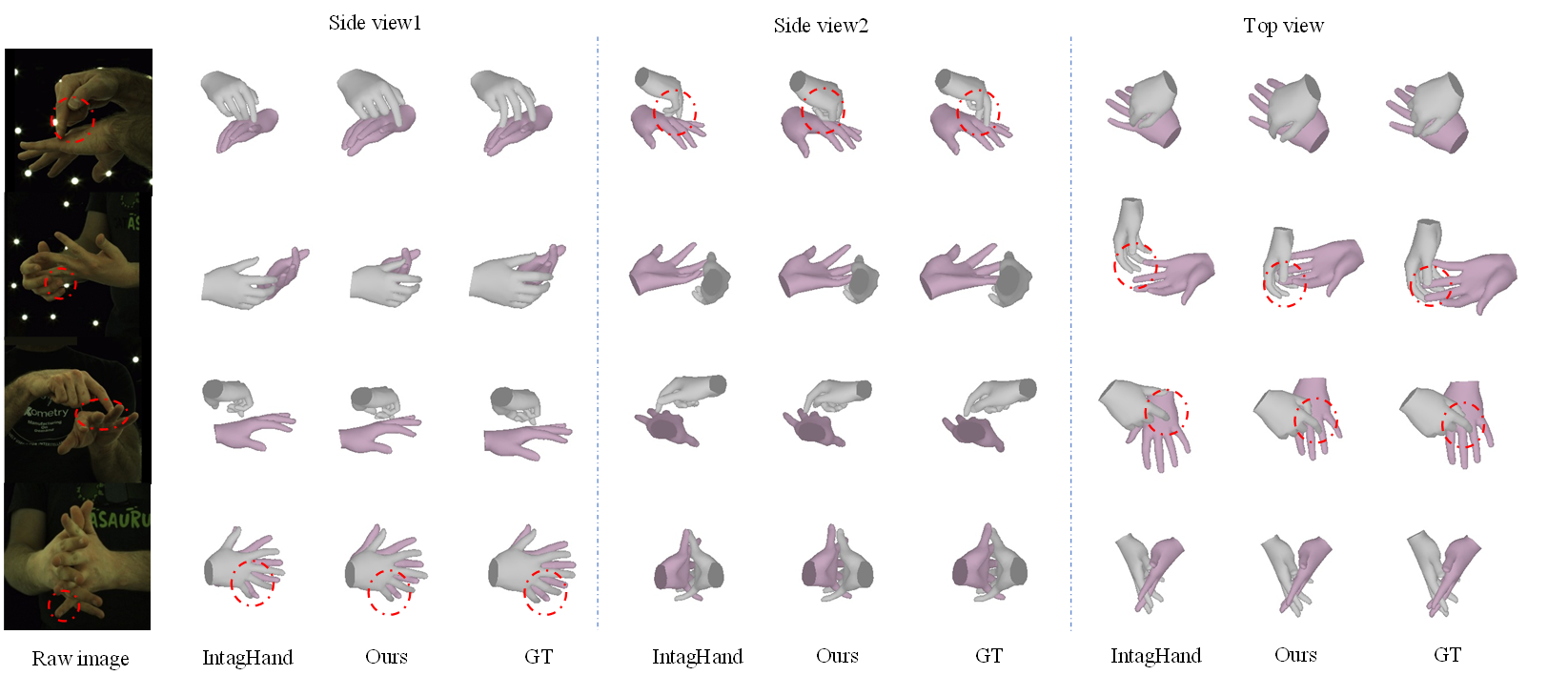}
	\caption{Qualitative comparison with our method and IntagHand~\cite{intaghand} on InterHand2.6M under a variety of viewpoints and different levels of inter-hand occlusion. Red circles are used to highlight the positions where our methods can generate better results. In the first row, our result can be even better than the ground truth, where the middle, ring, and little fingers of the right hand are curved.}
	\label{view}
\end{figure*}

\textbf{Optimization details.} As shown in Figure~\ref{anchor_pair}~a), The anchor pairs are built between the closest anchors on both hands, making the IH has more contact. As shown in Figure~\ref{anchor_pair}~b), to avoid abnormal anchors pairs, the pair can only be established when $\Bar{n_i^a}\cdot \Bar{n_j^a} < 0$, in which $\Bar{n^a}$ is the mesh face normal vector of the anchor. However, the IH attraction might cause a negative influence when the parts are in serious overlaps, as shown in Figure~\ref{anchor_pair}~c), there are conflicts between pairs, making the mesh hard to separate, the simple way to solve this problem is separating the hands at first so that we could have better anchor pairs. 

 \begin{eqnarray}
 \mathop{argmin}\limits_{\psi^r, \psi^l}( w_1\sum_{i=1}^{A_r}\sum_{j=1}^{A_l} L_{ij}^A + w_2 L_a + w_3 L_{adv} + w_4 L_p),
 \label{argmin_loss}
 \end{eqnarray}

 In our implementation, we optimize the loss function in Equation~\ref{argmin_loss} which is defined in the main paper in 215 iterations, we assign a larger weight $w_4$ for $L_{p}$ and a smaller weight $w_1$ for $L^A$ at the beginning to separate the hands, $w_1$ will increase while $w_4$ decrease during the optimization until 165th iteration. The anchor pairs will be rebuilt every 40 iterations to adapt to dynamically changing IH. The learning rate is set to 0.01 and will reduce after 20 no-loss-decaying iterations. Adam solver is utilized for optimization.

% \begin{table}
% \centering
% \caption{User study on natural rate. The higher the number, the more natural it is.}
% \begin{tabular}{cccc}
%    \toprule
%     With $\mathcal{D}$ & No $\mathcal{D}$ & Raw poses & Augmented poses \\
%    \midrule
%    81.25\% & 54.68\% & 90.82\% & 32.92\%\\
%    \bottomrule
% \end{tabular}
% \label{natural_test}
% \end{table}

\begin{figure*}[t]
	\centering
	\includegraphics[width=0.88\textwidth]{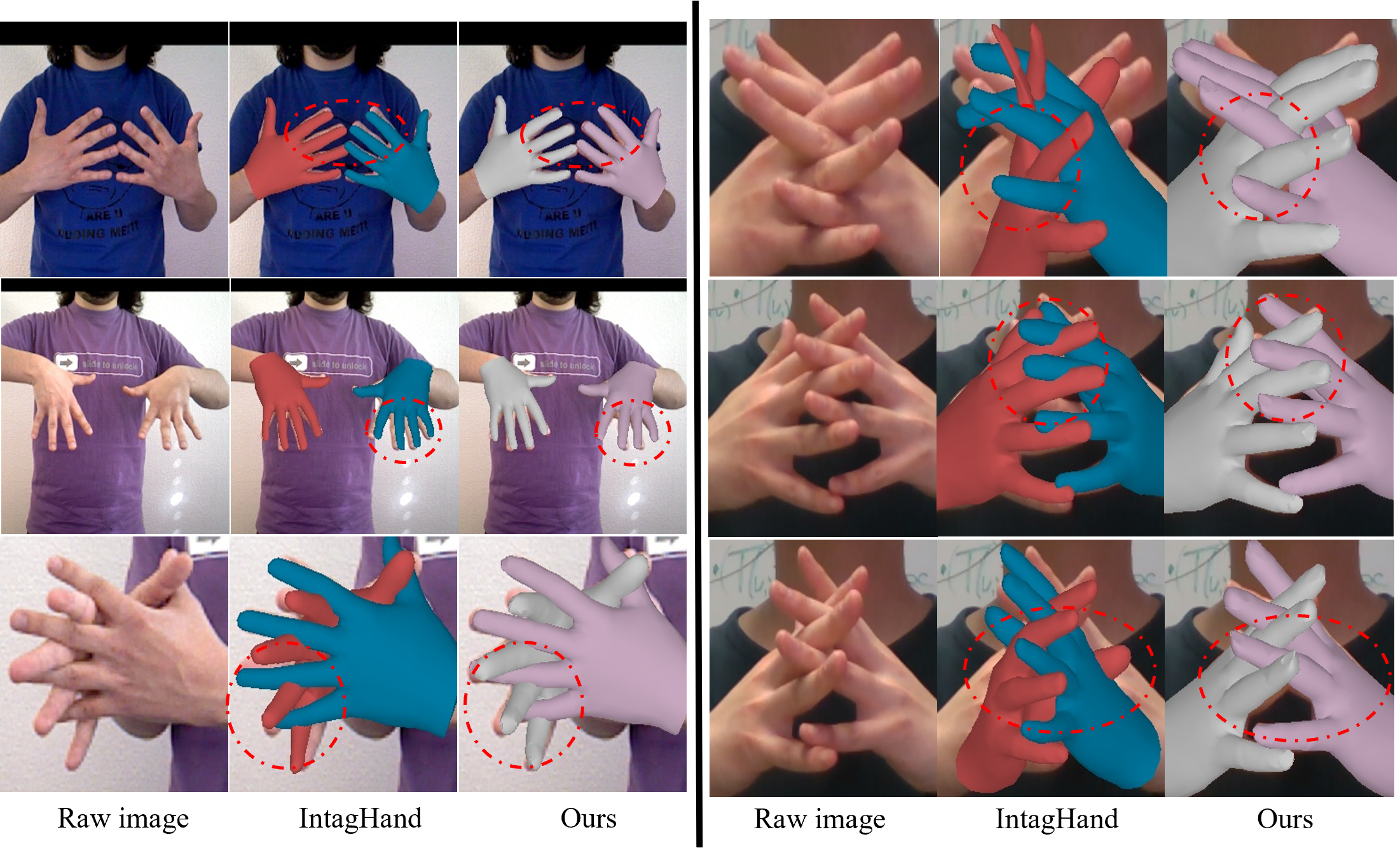}
	\caption{Qualitative comparison of the interacting hand pose estimation with our method and IntagHand~\cite{intaghand} on in-the-wild images. The first three columns display images from the Tzionas dataset~\cite{tzionas}, while the last three columns showcase images from the RGB2Hands dataset~\cite{rgb2hands}. Red circles are used to highlight the positions where our methods can generate better results. From the visualization results, we can clearly see that our model can generalize better for in-the-wild images.
	}
	\label{wild}
\end{figure*}

\section{More visualization results}
\label{section3}
\subsection{Results for different optimization components}
\textbf{Visualization of the effect of different components.} We define multiple optimization loss functions to get valid and natural IH poses. As shown in Figure~\ref{multiple_pic_const}, the ``Augmented Pose" is randomly augmented from the raw poses in IH2.6M, the joint poses are restricted according to Table~2 in the main paper. After being optimized by the full constraints, we get natural and non-interpenetration poses. Comparing Figure~\ref{multiple_pic_const}(b) and Figure~\ref{multiple_pic_const}(c), we can see that adopting anchors to make IH attraction has no significant differences from employing full vertices while reducing the time complexity. Furthermore, as demonstrated in Figure~\ref{multiple_pic_const}(d), the natural discriminator $\mathcal{D}$ could make the IH more natural, the {\textbf{natural}} poses are defined in the main paper, they not only conform to the anatomy but also frequently occur in daily life. Additionally, as shown in Figure~\ref{multiple_pic_const}(e), IH attraction enhances hand contact, which is hard to annotate in reality due to inter-occlusion.

\begin{figure}[h]
	\centering
	\includegraphics[width=0.45\textwidth]{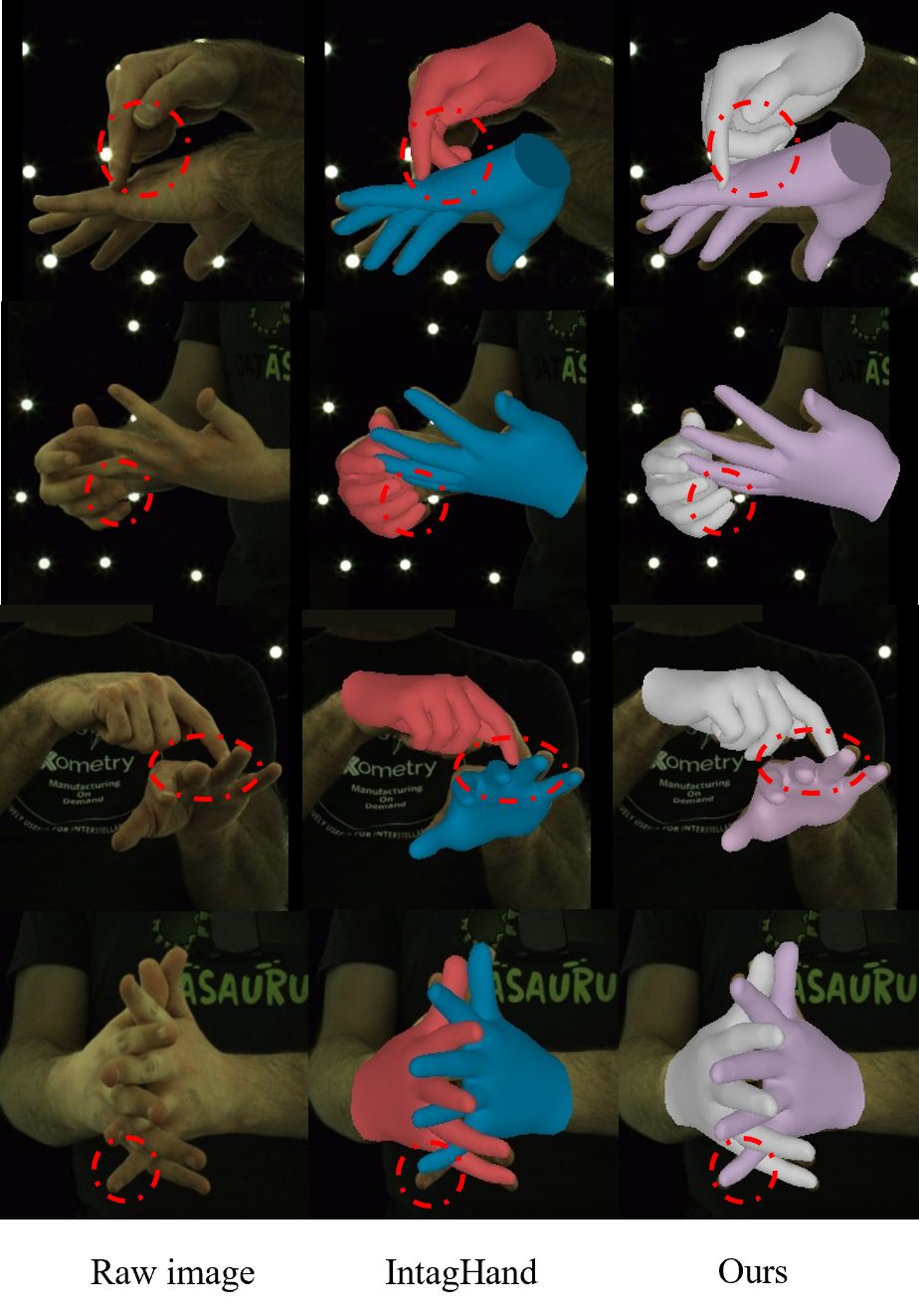}
	\caption{Qualitative comparison of the interacting hand pose estimation with our method and IntagHand~\cite{intaghand} on InterHand2.6M. Red circles are used to highlight the positions where our methods can generate better results.
	}
	\label{render}
\end{figure}

\begin{figure}[h]
    \centering
    \includegraphics[width=0.48\textwidth]{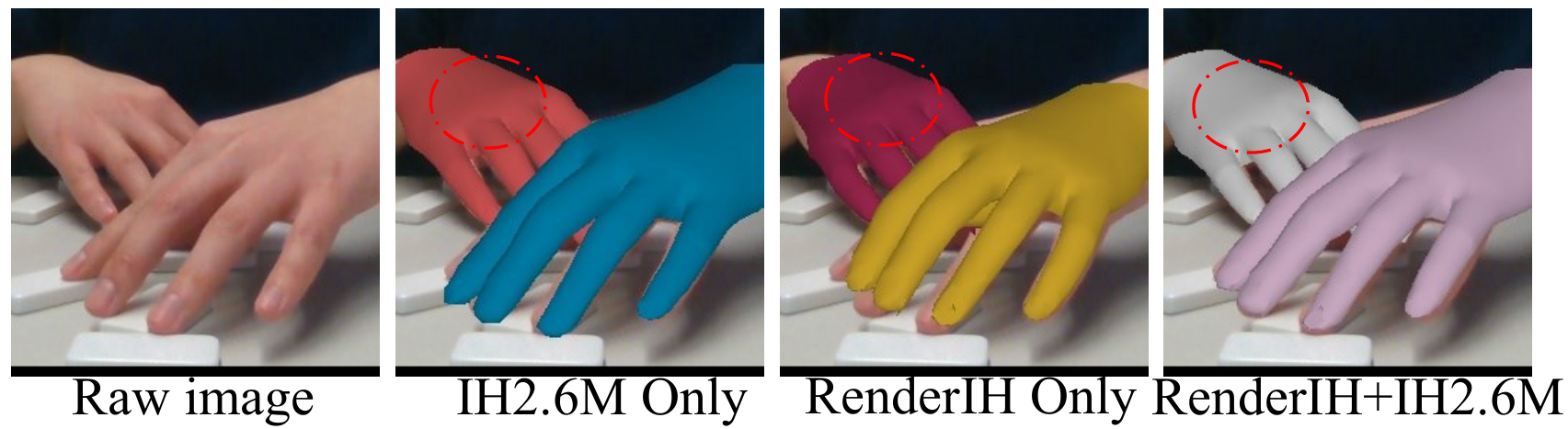}
	\caption{Qualitative comparison of our models trained on different training sets which test on in-the-wild images. 
	}
	\label{Comp}
\end{figure}

%\textbf{User study for naturalness}. Since the perceptions of "natural" may differ from human to human, We also conduct experiments to prove the discriminator's effect. We invited 20 persons with/without computer technical background, their ages are from 20 to 60, and the proportion of male to female was approximately 2:1. For each of them, we show 120 pictures (including 30 of augmented poses, 30 of optimized poses, 30 of optimized without discriminator, and 30 of raw poses from IH2.6M) of the IH poses, they are asked to determine whether the shown poses are natural, we count the NR (natural rate) of each category. The results are presented in Table~\ref{natural_test}, the ``Raw poses" are those from IH2.6M\cite{interhand26}, they are performed by humans and have high NR, however, some serious mesh-penetration caused by annotation mistakes might make the testers hardly to determine the ``natural". The ``Augmented poses" are augmented from the raw poses by assigning random rotation offsets to hand joints, they follow the joint limitation but have randomness, and some of them are in mesh penetration, the NR is low in this category. Optimizing the augmented poses without $\mathcal{D}$ solves the penetration, and the poses are valid, but the poses are not natural enough. It is clear that $\mathcal{D}$ improves the naturalness of the poses.

\begin{table}[h]
\centering
\small
\begin{tabular}{l|c}
    \toprule
  \textrm{Training~set}$\backslash$\textrm{Metrics} & {PAMPJPE/MPJPE/SMPJPE/MRRPE$\downarrow$}\\
        %\cline{1-2}
       %\textrm{Training~set}$\backslash$\textrm{Test~set} & IH2.6M \\
       % _{\rm{Training~set}}$\backslash$^{\rm{Test~set}} & Tzionas \\
        \midrule
                RenderIH & 13.50/47.73/49.42/32.08\\
                IH2.6M &  6.76/16.78/13.97/14.63\\
               IH2.6M+RenderIH & 5.79/15.78/12.16/14.15\\
        \bottomrule
        \end{tabular}
\caption{The comparison of training with or without our dataset and test on IH2.6M dataset. Wrist joint is used as root.}
\label{renderih}
%\vspace{-0.3cm}
\end{table}

\subsection{Qualitative results comparison}
\textbf{Comparison with IntagHand.} To better demonstrate the superiority of our data and method, we compare our result with the existing state-of-the-art method IntagHand~\cite{intaghand} (Their models is also trained on the combination of IH2.6M~\cite{interhand26} and synthetic images). Some qualitative comparisons with IntagHand are shown in Figure~\ref{render}. By directly projecting 3D hand mesh onto the image, we can see our result is closer to the pose in the raw image. Additionally, the results of these images from various views are also presented (see Figure~\ref{view}). In the first row of Figure~\ref{view}, our result can be even better than the ground truth, where the middle, ring, and little fingers of the right hand are curved. To further compare our generalization ability, we compare with IntagHand on in-the-wild images (see Figure~\ref{wild}). The results show that our method can clearly achieve less inter-penetration of two hands and more accurate finger interactions.

\textbf{Impact of synthetic data.} When only RenderIH is used for training, the performance is worse than when only IH2.6M is used, in part because the background variation in Tzionas is limited. The
trend can be seen in the qualitative result in Figure~\ref{Comp}. However, as a synthetic dataset, the function of our dataset is
to largely reduce the number of real data needed for training instead of replacing real data entirely.

\subsection{Quantitive results with wrist joint as root joint}
For convenient future comparison, we report our model's performance using wrist joint as root joint following common practice. As shown in Table~\ref{renderih}, the model trained on a mixture of RenderIH and IH2.6M demonstrates consistent improvement across all metrics compared to training on IH2.6M alone.

\section{Broader impacts and limitations}
\label{section4}

\textbf{Broader impacts.} In this paper, we introduce a synthetic 3D hand dataset, RenderIH, 
with accurate and diverse poses. Since there are no large-scale synthetic interacting hand datasets, RenderIH will be impactful for the community, due to its unprecedented scale, diversity, and rendering quality. Moreover, the dataset not only can be used to improve the generalization ability in real scenes but also can be used for domain adaptation.

\textbf{Limitations.}
The hyperparameters of pose optimization are chosen on the basis of experimental results, such as factor $k$ and $s$ in Interhand attraction and weights in the final optimization loss. In the future, we may set them as learnable parameters that can be automatically learned from data.

%%%%%%%%% REFERENCES
{\small
\bibliographystyle{ieee_fullname}
\bibliography{egbib}
}

% \end{document}
%\subsection{Video results}
% Since the InterHand2.6M~\cite{interhand26} dataset is collected in lab environment, the background is always black and the illumination is dark. When training only on InterHand2.6M, the network cannot perform well in real scenarios. In order to demonstrate generalization ability of RenderIH and our model, we collect two finger dancing videos from the Internet with rapid finger moving, complex actions and motion blur. More visualization results are presented on the complex Internet videos with temporal smoothing (see supplementary videos). Our results demonstrate accurate hand interaction even in the difficult scenarios and it benefits from the strong representation power of our TransHand model trained on RenderIH.
\end{document}